\documentclass{article}

\PassOptionsToPackage{numbers, compress}{natbib}

\usepackage[final]{nips_sty}

\usepackage{natbib}
\setcitestyle{numbers,square}
\usepackage{amsmath}
\usepackage{graphicx}
\usepackage{multirow}
\usepackage{multicol}
\usepackage[utf8]{inputenc} 
\usepackage[T1]{fontenc}    
\usepackage{url}            
\usepackage{booktabs}       
\usepackage{amsfonts}       
\usepackage{nicefrac}       
\usepackage{microtype}      
\usepackage{xcolor}         
\usepackage{makecell}
\usepackage{colortbl}
\usepackage{wrapfig}

\def\riou#1{$\text{RayIoU}_\text{#1}$}
\def\Mymth{OPUS}

\def\ie{\textit{i.e.}}
\def\etc{\textit{etc}}

\def\V#1{\rotatebox{90}{#1}}

\def\bevformer{BEVFormer~\cite{li2022bevformer}}
\def\renderocc{RenderOcc~\cite{pan2023renderocc}}
\def\bevdetocc{BEVDet-Occ~\cite{huang2021bevdet}}
\def\bevdetocceight{BEVDet-Occ (8f)~\cite{huang2021bevdet}}
\def\fbocc{FB-Occ (16f)~\cite{li2023fb}}
\def\sparseocceight{SparseOcc (8f)~\cite{liu2023fully}}
\def\sparseoccsixteen{SparseOcc (16f)~\cite{liu2023fully}}

\usepackage[pagebackref=false,breaklinks,colorlinks,citecolor=gray]{hyperref}

\usepackage[capitalize]{cleveref}
\crefname{section}{Sec.}{Secs.}
\Crefname{section}{Section}{Sections}
\crefname{table}{Tab.}{Tabs.}
\Crefname{table}{Table}{Tables}

\usepackage{subfigure}
\usepackage{parskip}
\usepackage{caption}
\title{\Mymth{}: Occupancy Prediction Using a Sparse Set}

\author{
    Jiabao Wang$^{1}$\thanks{Equal contribution. $^{\dagger}$Corresponding author.},
    Zhaojiang Liu$^{2*}$,
    Qiang Meng$^{3}$,
    Liujiang Yan$^{3}$,
    Ke Wang$^{3}$,
    Jie Yang$^{3}$, \\
    \textbf{Wei Liu$^{2}$,}
    \textbf{Qibin Hou$^{1, 4\dagger}$,}\
    \textbf{Ming-Ming Cheng$^{1, 4}$}\\
    $^1$VCIP, College of Computer Science, Nankai University \\
    $^2$Shanghai Jiao Tong University \quad
    $^3$KargoBot Inc.\quad
    $^4$NKIARI, Shenzhen Futian \\
    \url{https://github.com/jbwang1997/OPUS}
}

\begin{document}

\maketitle

\begin{abstract}

Occupancy prediction, aiming at predicting the occupancy status within voxelized 3D environment, is quickly gaining momentum within the autonomous driving community.
Mainstream occupancy prediction works first discretize the 3D environment into voxels, then perform classification on such dense grids. However, inspection on sample data reveals that the vast majority of voxels is unoccupied. Performing classification on these empty voxels demands suboptimal computation resource allocation, and reducing such empty voxels necessitates complex algorithm designs.
To this end, we present a novel perspective on the occupancy prediction task: formulating it as a streamlined set prediction paradigm without the need for explicit space modeling or complex sparsification procedures.
Our proposed framework, called \Mymth{}, utilizes a transformer encoder-decoder architecture to simultaneously predict occupied locations and classes using a set of learnable queries.
Firstly, we employ the Chamfer distance loss to scale the set-to-set comparison problem to unprecedented magnitudes, making training such model end-to-end a reality.
Subsequently, semantic classes are adaptively assigned using nearest neighbor search based on the learned locations.
In addition, \Mymth{} incorporates a suite of non-trivial strategies to enhance model performance, including coarse-to-fine learning, consistent point sampling, and adaptive re-weighting, \etc.
Finally, compared with current state-of-the-art methods, our lightest model achieves superior RayIoU on the Occ3D-nuScenes dataset at near $2\times$ FPS, while our heaviest model surpasses previous best results by 6.1 RayIoU.

\end{abstract}
\section{Introduction}

Compared with well-established 
box representations~\cite{huang2021bevdet,liu2022petr,liu2023sparsebev,wang2024towards,meng2024small,zhu2023curricular,shi2022srcn3d}, voxel based occupancy~\cite{li2023fb,tang2024sparseocc,jia2023occupancydetr,tian2024occ3d, shi2024effocc} can provide finer geometry and semantic information for the surrounding scene. For example, it is not straightforward to use bounding boxes to describe vehicles with doors open or cranes with outriggers deployed. While occupancy can naturally describe such uncommon shapes. Thus occupancy prediction is quickly gaining traction in the autonomous driving community.


Recent approaches~\cite{cao2022monoscene,zhang2023occformer,huang2023tri,ma2023cotr,li2023fb, shi2024panossc} to the task predominantly rely on dense data representation, 
with a direct one-to-one correspondence between feature points and physical voxels.
It has come to our attention that the vast majority of physical voxels is empty. For instance, in SemanticKITTI~\cite{behley2019semantickitti}, approximately 67\% of all voxels are empty, while in Occ3D-nuScenes~\cite{tian2024occ3d}, this proportion exceeds 90\%.
Such sparse nature of occupancy data renders the direct dense representation undeniably inefficient, as majority of the computation is allocated towards empty voxels.
Alternative sparse latent representations have been explored to alleviate such inefficiency, such as the Tri-Perspective View representation~\cite{tang2024sparseocc, huang2023tri} or reduced solution spaces~\cite{liu2023fully,jia2023occupancydetr}, leading to notably reduced computational costs.
However, these approaches still treat occupancy prediction as a classification problem at specific locations, necessitating complex intermediate designs and explicit modeling of 3D spaces.

\begin{figure}
    \centering
    \includegraphics[trim={0cm 0cm 0cm 0cm},clip, width=0.95\textwidth]{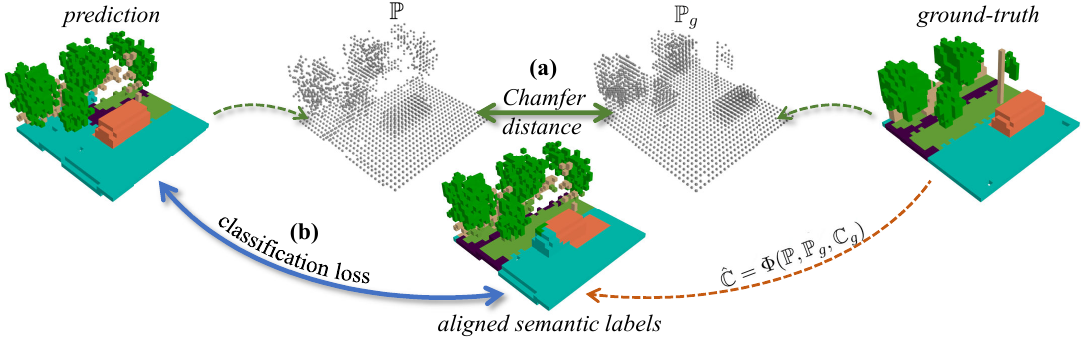}
    \caption{
        The occupancy prediction is approached as a set prediction problem.
        For each scene, we predict a set of point positions $\mathbb{P}$ and a set of the corresponding semantic classes $\mathbb{C}$.        
        With the ground-truth set of occupied voxel positions $\mathbb{P}_g$ and classes $\mathbb{C}_g$, we decouple the set-to-set matching task into two distinct components:
        (a) Enforcing similarity in the point distributions of $\mathbb{P}$ and $\mathbb{P}_g$ using the Chamfer distance.
        (b) Aligning the predicted classes $\mathbb{C}$ with the ground-truths $\hat{\mathbb{C}} = \Phi(\mathbb{P}, \mathbb{P}_g, \mathbb{C}_g)$, where $\Phi$ generates a set of classes for points $\mathbb{P}$ based on those of the nearest ground-truth points.
    }\label{fig:intro}
    \vspace{-10pt}
\end{figure}

In this work, we instead formulate the task as a direct set prediction problem, where we regress occupied locations and classify corresponding semantic labels in parallel.
Our proposed framework termed \Mymth{} leverages a transformer encoder-decoder architecture featuring:
(1) an image encoder to extract 2D features from multi-view images;
(2) a set of learnable queries to predict occupied locations and semantic  classes;
(3) a sparse decoder to update query features with correlated image features.
Our \Mymth{} eliminates the need for explicit space modeling or complex sparsification procedures, offering a streamlined and elegant end-to-end solution. 
However, a key challenge lies in matching predictions with ground-truths, especially given the unordered nature of predicted results.
We argue that the Hungarian algorithm~\cite{kuhn1955hungarian}, although widely adopted in the DETR families~\cite{carion2020end,zhu2020deformable, meng2021conditional, liu2022dab, li2022dn, wang2022anchor}, is not suitable for this task. Having a $O(n^3)$ time complexity and a $O(n^2)$ space complexity, the Hungarian algorithm is unable to handle a substantial number of voxels. In our experiments,
associating two sets with 10K points each, the Hungarian algorithm consumes approximately 24 seconds and 2,304Mb of GPU memory on a 80G A100 GPU.
In reality, the voxel number can go up to $\sim$70K in the Occ3D-nuScenes~\cite{tian2024occ3d} dataset. Thus directly applying the Hungrian algorithm for set-to-set matching is infeasible in the occupancy prediction context.


But is accurate one-to-one association truly necessary for occupancy prediction? We recognize that the goal of one-to-one correspondence between prediction results and ground-truth annotation is to obtain supervision signals, essentially
 complete, precise \textit{point locations}, and accurate \textit{point classes}.
The heavylifting of one-to-one association can be entirely avoided if we can obtain such supervision signals elsewhere.
Therefore, we propose to decouple the occupancy prediction task into two parallel subtasks, as illustrated in \cref{fig:intro}.
The first task obtains supervision on point locations by aligning predicted point distributions with ground-truths, a task achievable through the Chamfer distance loss, a well-established technique for point clouds~\cite{fan2017point,mersch2022self}.
The second task obtains supervision on point classes by assigning semantic labels to predicted points.
This is accomplished by assigning each point the class of its nearest neighbor in the ground-truths.
It's noteworthy that all operations involved can be executed in parallel and are highly efficient on GPU devices.
As a result, a single matching in Occ3D-nuScenes can be processed within milliseconds, with negligible memory consumption.
With a time complexity of $O(n^2)$ and space complexity of $O(n)$, 
our formulation breaks the ground for large-scale training for the occupancy prediction models.

In addition, we propose several strategies to further boost the performance of occupancy prediction in our end-to-end sparse formulation, including coarse-to-fine learning, consistent point sampling, and adaptive loss re-weighting.
On Occ3D-nuScenes, all our model variants easily surpass all prior work, verifying the efficacy and effectiveness of the proposed method. 
Especially, our most lightweight model achieves a 3.3 absolute \riou{} improvement compared with SparseOcc~\cite{liu2023fully} while operating more than $2\times$ faster.
The heaviest configuration ultimately achieves a \riou{} of 41.2, establishing a new upper bound with a 14\% advantage.
Our contributions are summarized as follows:


\begin{itemize}
    \item To the best of our knowledge, for the first time, we view the occupancy prediction as a direct set prediction problem, facilitating end-to-end training of the sparse framework. 
    \item Several non-trivial strategies, including coarse-to-fine learning, consistent point sampling, and adaptive re-weighting, are further introduced for boosting the performance of OPUS. 
    \item Extensive experiments on Occ3D-nuScenes reveal that \Mymth{} can outperform state-of-the-art methods in terms of RayIoU results, while maintain a real-time inference speed.
\end{itemize}
\section{Related work}

\subsection{Occupancy prediction}
Occupancy prediction entails determining the occupancy status and class
of each voxel within a 3D space.
This task has recently become a foundational perception task in autonomous driving and raises great interests from both academic and industrial communities.
Conventional methods~\cite{cao2022monoscene,zhang2023occformer,huang2023tri,ma2023cotr,li2023fb,wang2023openoccupancy,tian2024occ3d, cao2024vision} typically employ the continuous and dense feature representation, which, however, suffer from computational redundancy due to the inherent sparsity of occupancy data.
In addressing this issue,  \citet{tang2024sparseocc} compresses the dense feature using the Tri-Perspective View representation for model efficiency.
Recently, several transformer-based approaches~\cite{liu2023fully,jia2023occupancydetr,li2023voxformer} with sparse queries have emerged. 
For example, OccupancyDETR~\cite{jia2023occupancydetr} conducts object detection followed by assigning each object with one query for occupancy completion.
VoxFormer~\cite{li2023voxformer} generates 3D voxels from a set of sparse queries, corresponding to occupied locations identified through a pre-task of depth estimation.
Meanwhile, SparseOcc~\cite{liu2023fully} employs a series of sparse voxel decoders to filter out empty grids and predict occupied statuses of retained voxels in each stage.
While these approaches have succeeded in reducing computational costs, they often necessitate multi-stage processes and intricate space modeling.
In contrast, our method directly applies sparse queries to regress the occupied locations without pre-defined locations, facilitating an elegant and end-to-end occupancy prediction.

\subsection{Set prediction with transformers}
The concept of directly predicting sets with Transformers was initially introduced by DETR~\cite{carion2020end}, where a set of sparse queries generates unordered detection results with feature and object interactions.
By viewing the object detection as a direct set prediction problem, DETR eliminates the need for complex post-processing, enabling end-to-end performance.
Following DETR, numerous variants~\cite{zhu2020deformable,meng2021conditional,liu2022dab,li2022dn,wang2022anchor,zhang2022dino,sun2021rethinking} have been proposed for performance improvements and efficient training.
The effectiveness of the sparse-query-based paradigm has also been validated in 3D object detection~\cite{wang2022detr3d,liu2022petr,lin2022sparse4d,lin2023sparse4d,liu2023sparsebev,wang2023exploring}, where 3D information is encoded into the queries.
For example, DETR3D~\cite{wang2022detr3d} employs a sparse set of 3D object queries to index 2D features, linking 3D positions to multi-view images using camera transformation matrices.
PETR~\cite{liu2022petr} generates 3D position-aware features by encoding 3D position embedding into 2D image features, enabling queries to directly aggregate features without the 3D-to-2D projection.
Sparse4D~\cite{lin2022sparse4d} further advances sparse 3D object detection by refining detection results with spatial-temporal feature fusion.
Despite of the great success, set prediction with Transformers remains restricted primarily to object detection, where the query number are typically small due to the limited object number in a scene.
Extending this approach to occupancy prediction poses a big challenge due to the substantially larger number of queries required.

\section{Methodology}

In this part, we first recap current query-based sparsification approaches for occupancy prediction in \cref{sec:revisiting}.
Then, \cref{sec:our_sparse} describes our formulation that views the task as a direct set prediction problem.
Finally, we detail the proposed \Mymth{} framework in \cref{sec:otr}.


\subsection{Revisiting query-based occupancy sparsification}\label{sec:revisiting}

Transformers with sparse queries offer a promising avenue for tackling the inherent sparsity in occupancy representation. 
A notable approach to reduce the number of queries is allocating each query to a patch of voxels rather than a single voxel, as presented in PETRv2~\cite{liu2023petrv2}.
However, this method still generates a dense prediction of the 3D space, thus failing to efficiently address the redundancy issue.
Alternatively, VoxFormer~\cite{li2023voxformer} and SparseOcc~\cite{liu2023fully} allocate sparse queries exclusively to occupied voxels. 
VoxFormer employs a depth estimation module to identify potentially occupied voxels, while SparseOcc utilizes multiple stages to progressively filter out empty regions. 
Nonetheless, their sparsification processes rely on accurately recognizing the occupancy status of voxels and therefore suffer from the cumulative errors.
Moreover, their pipelines necessitate intricate intermediate descriptions of the 3D space, hindering seamless end-to-end operation.

The dilemmas of current  approaches significantly stem from treating the task as a classification problem, where each query is confined to a specific physical region for classifying the semantic labels. 
This constraint severely limits query flexibility, preventing adaptive focus on suitable areas. 
To address this, we propose to remove this restriction by allowing each query to autonomously determine its relevant area. 
In the end, we view occupancy prediction as a direct set prediction problem, where each query predicts point positions and semantic classes, simultaneously.


\subsection{A set prediction problem}\label{sec:our_sparse}
At the core of our work lies the conceptualization of occupancy prediction as a set prediction task.
We denote the $V_g$ occupied voxels in the ground-truth as $\{\mathbb{P}_g, \mathbb{C}_g\}$, where $|\mathbb{P}_g| = |\mathbb{C}_g| = V_g$.
For each entry in $\{\mathbf{p}_g, \mathbf{c}_g\}\in \{\mathbb{P}_g, \mathbb{C}_g\}$, $\mathbf{p}_g$ represents the 3D coordinates of a voxel center, while $\mathbf{c}_g$ stores the semantic class of the corresponding voxel.
Given the predictions $\{\mathbb{P}, \mathbb{C}\}$ of $V$ points, our primary challenge is to devise an effective strategy for set-to-set matching.
In other words, we must determine how to supervise the training of unordered predictions with the ground-truth data.
One alternative is to adopt the Hungarian algorithm.
However, our previous discussions and experiments in the appendix reveal its scalability limitations.
Rather than pursuing one-to-one associations between the predictions and ground-truths, we recognize the matching essentially aims at accurate locations and classes in predictions.
This motivates us to decouple the task into two parallel objectives:
(1) Encouraging the predicted locations to be precise and comprehensive.
(2) Ensuring the predicted points are assigned with proper semantic classes from the ground-truth labels.

The first objective focuses on aligning distributions between predicted and ground-truth points, a task achievable through the Chamfer distance loss which is well-proved in the field of point clouds~\cite{fan2017point,mersch2022self,khurana2023point,yang2024visual}:
\begin{equation}
    \small
    \text{CD}(\mathbb{P}, \mathbb{P}_g) = \frac{1}{|\mathbb{P}|}\sum\limits_{\mathbf{p} \in \mathbb{P}} D(\mathbf{p}, \mathbb{P}_g) + \frac{1}{|\mathbb{P}_g|}\sum\limits_{\mathbf{p}_g \in \mathbb{P}_g} D(\mathbf{p}_g, \mathbb{P}), \text{ where } D(\mathbf{x}, \mathbb{Y}) = \min_{\mathbf{y} \in \mathbb{Y}}||\mathbf{x} - \mathbf{y}||_1.
    \label{eq:label_loc}
\end{equation}
Minimizing Chamfer distance leads to similar distributions of predictions and ground-truths, enabling direct learning of occupied voxels without necessitating knowledge of their orders.

Concerning the second objective, although direct comparison between $\mathbb{C}$ and $\mathbb{C}_g$ is invalid due to their correspondence to different locations, we can leverage the spatial locality properties of voxels to find a proxy.
Nearby points belonging to the same object usually carry the same semantic labels, 
thus we propose assigning each predicted point the class of its nearest neighbor voxel in the ground-truth: 
\begin{equation}
    \small
    \{\hat{\mathbb{C}}, \hat{\mathbb{P}}\} = \left\{
        {\arg\min}_{\{\mathbf{c}_g, \mathbf{p}_g\}\in \{\mathbb{C}_g, \mathbb{P}_g\}} \|\mathbf{p}_g - \mathbf{p}\|_2, \quad \mathbf{p} \in \mathbb{P}
    \right\}.
    \label{eq:label_cls}
\end{equation}
Here, $\hat{\mathbb{C}}$ is the updated classes that are prepared to supervise the learning of the predicted $\mathbb{C}$.

It's noteworthy that computations of both \cref{eq:label_loc} and \cref{eq:label_cls} can be executed efficiently and in parallel on GPU devices.
As a result, a single matching can be swiftly processed within milliseconds, enabling feasibility of the large-scale training by viewing the occupancy prediction task as a direct set prediction problem.
Next, we delve into the specifics of the proposed \Mymth{} framework.

\subsection{Details of \Mymth{}}\label{sec:otr}

\begin{figure*}[!t]
    \centering
    \includegraphics[width=0.97\textwidth]{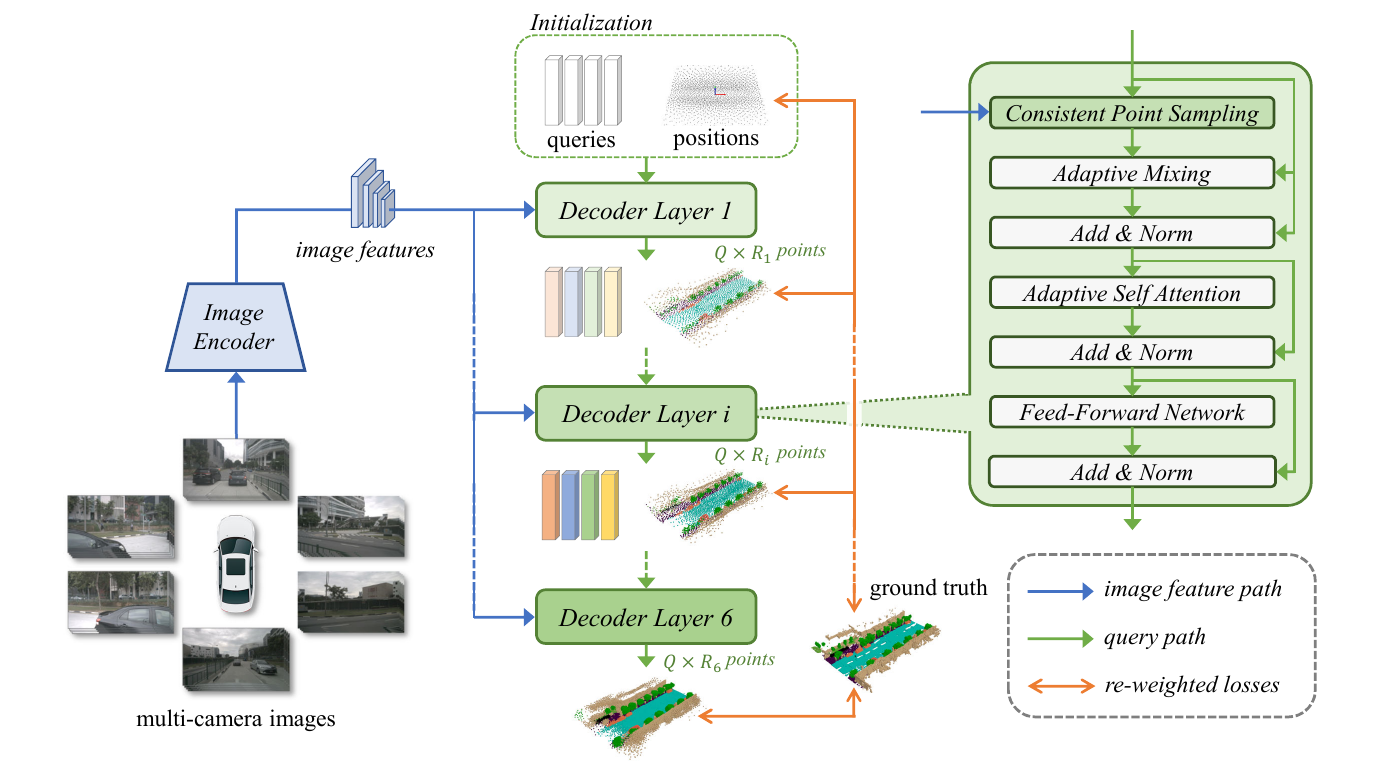}
    \caption{
    \Mymth{} leverages a transformer encoder-decoder architecture comprising:
    (1) An image encoder to extract 2D features from multi-view images.    
    (2) A series of decoders to refine the queries with image features, which are correlated via the \textit{consistent point sampling} module.
    (3) A set of learnable queries to predict locations and classes of occupancy points.
    Each query obeys a \textit{coarse-to-fine} rule, progressively increasing the number of predicted points.
    In the end, the entire model is trained end-to-end using our \textit{adaptively re-weighted} set-to-set losses.
    }
    \label{fig:structure}
    \vspace{-10pt}
\end{figure*}

This part describes \Mymth{} framework, as illustrated in \cref{fig:structure}.
Initially, image features are extracted from multi-view images. And a set of learnable queries $\mathbb{Q}$, point positions $\mathbb{P}$, and scores $\mathbb{C}$ are initialized.
Subsequently, these query features and prediction outcomes are fed into a sequence of decoders, undergoing iterative refinement through correlation with image features.
At each stage, predicted positions and scores are supervised by the ground-truths, facilitating end-to-end training for the entire framework.
It can be observed that our most important structure is the sequence of multiple decoders.
Therefore, we next provide a detailed description to the inputs/outputs of the decoders and how features are aggregated and updated within the decoders.

\textbf{Notations.}
Denote the set of learnable queries, point positions, and point scores as $\{\mathbb{Q}_0, \mathbb{P}_0, \mathbb{C}_0\}$ 
before feeding into decoders, and as $\{\mathbb{Q}_{i}, \mathbb{P}_{i}, \mathbb{C}_{i}\}$ for the outcomes of the $i$-th decoder.
The length of these sets is all $Q$, which corresponds to the number of queries.
Each query feature $\mathbf{q}_i\in \mathbb{Q}_i, i \in \{0, 1, \cdots, 6\}$ has a channel size $C$, set to 256 in our implementation.
To reduce the number of queries, which is a bottleneck for model efficiency, each query $q_i$ predicts $R_i$ points rather than a single one.
Consequently, $\mathbf{p}_i\in \mathbb{P}_i$ and $\mathbf{c}_i \in \mathbb{C}_i$ have shapes of $Q \times R_i \times 3$ and $Q \times R_i \times N$, respectively.
Here, $N$ represents the number of semantic classes.

\textbf{Coarse-to-fine prediction.}
High-level semantic information can be difficult to predict accurately from just low-level features.
Therefore, instead of attempting to predict occupancy for the entire 3D environment, we allow the model to predict "sparse" occupancy results in early stages, as shown in \cref{fig:structure}.
To achieve this, we follow a coarse-to-fine strategy, gradually increasing the number of points generated from one query.
In other words, we always have $R_{i-1}\leq R_i$ for $i\in \{1, 2, \cdots, 6\}$.

It's noteworthy that the Chamfer distance has another advantage over the Hungarian algorithm here:
even when the number of predictions is smaller than that of the ground-truths, the assignment won't collapse into a local shape of the ground-truths.
This is because the Hungarian algorithm could assign the predictions to any subset of the ground-truths due to its lack of  distribution constraints.
In contrast, the Chamfer distance maintains a global perspective, considering the overall distribution of points rather than enforcing a strict one-to-one correspondence. 
This ensures that the predicted points are more evenly distributed and representative of the actual 3D environment, even when fewer in number.

\textbf{Details of the decoder.}
Our decoder is analogous to that in SparseBEV~\cite{liu2023sparsebev}, a performant and sparse object detector.
For a given query $\mathbf{q}_{i-1}\in \mathbb{Q}_{i-1}$ and its corresponding point locations $\mathbf{p}_{i-1} \in \mathbb{P}_{i-1}$, the $i$-th decoder first aggregates image features through a consistent point sampling, a new scheme elaborated in our subsequent discussion.
Subsequently, the query feature is updated into $\mathbf{q}_{i}$ with  the adaptive mixing of image and query features, along with the self-attention among all queries, mirroring operations in SparseBEV.
In the end, a prediction module, comprising only Linear, LayerNorm, and ReLU layers, generates the semantic classes $\mathbf{c}_{i}$ (size $R_i \times N$) and the position offsets $\Delta \mathbf{p}_{i}$ (size $R_i \times 3$).
As the $\Delta \mathbf{p}_{i}$ cannot be directly added to $p_{i-1}$ due to dimension misalignment, we first compute the mean of $\mathbf{p}_{i-1}$ along the first dimension and then duplicate the results by $R_i$ times into $\mathbf{\bar{p}}_{i-1}$. 
The final position $\mathbf{p}_i$ is computed as $\mathbf{p}_i = \mathbf{\bar p}_{i-1} + \Delta \mathbf{p}_{i}$.

\textbf{Consistent point sampling.}
The feature sampling method utilized in SparseBEV is not applicable for our method as it is specifically designed for detection inputs.
Therefore, we propose a novel process of Consistent Point Sampling (CPS), aiming at sampling 3D points and aggregating features from $M$ image features.
Given input $\{\mathbf{q}, \mathbf{p}\}\in \{\mathbb{Q},\mathbb{P}\}$, we sample $S$ points and find their respective coordinates in the $m$-th image feature by the following equation:
\begin{equation}
    \mathbf{c}_m = \mathbf{T_m} \mathbf{r}, \text{ where } \mathbf{r} = \mathbf{m}_p + \phi(\mathbf{q}) \cdot \mathbf{\sigma}_p,
    \label{eq:method_cps}
\end{equation}
where $\mathbf{T_m}$ represents the projection matrix from current 3D space into the $m$-th image's coordinates.
$\phi(\mathbf{q})$ generates $S$ 3D points from the query feature $\mathbf{q}$ using a linear layer.
$\mathbf{m}_p$ and $\mathbf{\sigma}_p$ denote the mean and standard deviation, respectively, of the $R$ points in $\mathbf{p}$.
It's worthy to note that we re-weight the predicted offsets $\phi(\mathbf{q})$ with the standard deviation $\mathbf{\sigma}_p$ to inherent the dispersion degree from previous predictions.
In essence, we tend to sample more aggressively if the input $\mathbf{p}$ contains diverse points, and sample points in a narrower range otherwise.
This operation can evidently enhance the prediction performance, as demonstrated in our experiments.

Not all coordinates in $\mathbf{c}_m$ are feasible since the sampled points might not be visible within the corresponding camera.
Therefore, we generate a mask set $\mathbb{V}_m$ where the $s$-th value is 1 if $\mathbf{c}_{s,m}$ is valid and 0 otherwise, for $s\in\{1,2,\cdots, S\}$ and $m\in\{1,2,\cdots, M\}$.
Next, we aggregate information from image features $\{F_m\}_1^M$ for the later adaptively mixing stage. 
Specifically, we have 
\begin{equation}
    \small
    f_s=\frac{1}{\sum_{m=1}^M |\mathbb{V}_m|}
    \sum_{s=1}^S \sum_{m=1}^M  w_{s,m} \cdot v_{s,m}\cdot \mathcal{B}(F_m, \mathbf{c}_{s,m}),
\end{equation}
where $v_{s,m}$ denotes the $s$-th element in $\mathbb{V}_m$ and $\mathbf{c}_{s,m}$ is the coordinates of the $s$-th point $\mathbf{r}_s$ mapped into the $m$-th image feature.
The operation $\mathcal{B}$ refers to the bilinear interpolation.
$w_{s, m}$ is the weight for the $\mathbf{r}_s$ on the $m$-th image feature, generated from the query feature $\mathbf{q}$ by linear transformation.

\textbf{The training loss with adaptively re-weighting.}
The training object of our framework is to supervise the learning of $\{\mathbb{P}_i,  \mathbb{C}_i\}_{i=1}^6$ with the ground-truth $\{\mathbb{P}_g,  \mathbb{C}_g\}$.
Point positions can be trained with \cref{eq:label_loc}.
However, the original Chamfer distance loss focuses on the overall similarity of point distributions, neglecting whether each individual is good enough.
This leads to unsatisfactory performance, as observed in our experiments.
To cope with this issue, we employ a simple but effective re-weighting strategy to emphasize erroneous points, and modify the Chamfer distance loss as follows:
\begin{equation}
    \small
    \begin{split}
    \text{CD}_R(\mathbb{P}, \mathbb{P}_g) &= \frac{1}{|\mathbb{P}|}\sum\limits_{\mathbf{p} \in \mathbb{P}} D_R(\mathbf{p}, \mathbb{P}_g) + \frac{1}{|\mathbb{P}_g|}\sum\limits_{\mathbf{p}_g \in \mathbb{P}_g} D_R(\mathbf{p}_g, \mathbb{P}),  \\
    \text{ where } & D_R(\mathbf{x}, \mathbb{Y}) = W(d)\cdot d \text{ with } d = \min_{\mathbf{y} \in \mathbb{Y}}||\mathbf{x} - \mathbf{y}||_1.
    \end{split}
    \label{eq:label_loc1} 
\end{equation}
Here, $W(d)$ is the re-weighting function penalizing points with large distance to the closest ground-truths.
In our implementation, we use a step function of $W(d)$ being 5 if $d \geq 0.2$ and 1 otherwise.

For the classification, we first generate the target classes $\hat{\mathbb{C}}_i$ for $\mathbb{C}_i$ using \cref{eq:label_cls}.
Subsequently, the semantic classes can be trained with the conventional classification losses.
In our implementation, we adopt the focal loss~\cite{lin2017focal} with mannually searched weights on different categories and denote the modified loss as FocalLoss$_R$.
In the end, the training objective of the proposed \Mymth{} becomes
\begin{equation}
    \small
    \label{eq:final}
    L_\text{\Mymth{}} = \text{CD}_R(\mathbb{P}_0, \mathbb{P}_g) + \sum\limits_{i=1}^6(\text{CD}_R(\mathbb{P}_i, \mathbb{P}_g) + \text{FocalLoss}_R(\mathbb{C}_i, \hat{\mathbb{C}}_i)),
\end{equation}
where $\text{CD}_R(\mathbb{P}_0, \mathbb{P}_g)$ explicitly encourages initial points $\mathbb{P}_0$ to capture a general pattern of the dataset.

\section{Experiments}


\subsection{Experimental setup}\label{sec:exp_setup}

\textbf{Dataset and metrics.}
All models are evaluated on the Occ3D-nuScenes~\cite{tian2024occ3d} dataset, which provides occupancy labels for 18 classes (1 free class and 17 semantic classes) on the large-scale nuScenes~\cite{caesar2020nuscenes} benchmark. 
Out of the 1,000 labeled driving scenes, 750/150/150 are used for training/validation/testing, respectively.
The commonly used mIoU metric is utilized for evaluation. 
Recently, SparseOcc~\cite{liu2023fully} points that that overestimation can easily hack the mIoU metric and proposes \riou{} as a remedy. 
Therefore, following their work, we also report the \riou{} results under different distance thresholds at 1, 2, and 4 meters, denoted as \riou{1m}, \riou{2m}, and \riou{4m}, respectively. 
The final \riou{} score is the average of these three values.


\textbf{Implementation details.}
Following previous works~\cite{liu2023fully,li2023fb,huang2021bevdet}, we resize images to $704 \times 256$ and extract features using a ResNet50~\cite{he2016deep} backbone.
We denote a series of models as \Mymth{}-T, \Mymth{}-S, \Mymth{}-M and \Mymth{}-L, with 0.6K, 1.2K, 2.4K and 4.8K queries, respectively.
In each model, all queries predict an equal number of points, totalling 76.8K points in the final stage.
The sampling number in our CPS is 4 for \Mymth{}-T and 2 for other models.
Please refer to \cref{sec:config} for more details of different models.
All models are trained on 8 nvidia 4090 GPUs with a batch size of 8 using the AdamW~\cite{loshchilov2018decoupled} optimizer.
The learning rate warms up to $2e^{-4}$ in the first 500 iterations and then decays with a Cosine Annealing~\cite{loshchilov2016sgdr} scheme. 
Unless otherwise stated, models in main results are trained for 100 epochs and those in the ablation study are trained for 12 epochs.


\subsection{Main results}\label{sec:exp_main}

\begin{table}[t]
    \small
    \caption{
        Occupancy prediction performance on Occ3D-nuScenes~\cite{tian2024occ3d}. 
        "8f" and "16f" denote models fusing temporal information from 8 or 16 frames, respectively.
        Baseline results are directly copied from their corresponding papers or the SparseOcc~\cite{liu2023fully}. 
        FPS results are measured on an A100 GPU.
    }
    \label{tab:main}
    \centering
    \scalebox{0.85}{
    \begin{tabular}{l|ccc|ccc|>{\columncolor{lightgray!20}}c|>{\columncolor{lightgray!20}}c}
    \toprule[0.9pt]
    Methods             & Backbone & Image Size      & mIoU & \riou{1m} & \riou{2m} & \riou{4m} & \riou{} & FPS \\ \midrule
    \renderocc{}        & Swin-B   & $1408\times512$ & 24.5 & 13.4      & 19.6      & 25.5      & 19.5    & - \\
    \bevformer{}        & R101     & $1600\times900$ & 39.3 & 26.1      & 32.9      & 38.0      & 32.4    & 3.0 \\
    \bevdetocc{}        & R50      & $704\times256$  & 36.1 & 23.6      & 30.0      & 35.1      & 29.6    & 2.6 \\
    \bevdetocceight{}   & R50      & $704\times384$  & 39.3 & 26.6      & 33.1      & 38.2      & 32.6    & 0.8 \\
    \fbocc{}            & R50      & $704\times256$  & 39.1 & 26.7      & 34.1      & 39.7      & 33.5    & 10.3 \\
    \sparseocceight{}   & R50      & $704\times256$  & -    & 28.0      & 34.7      & 39.4      & 34.0    & 17.3 \\
    \sparseoccsixteen{} & R50      & $704\times256$  & 30.6 & 29.1      & 35.8      & 40.3      & 35.1    & 12.5 \\ \midrule
    \Mymth{-T (8f)}     & R50      & $704\times256$  & 33.2 & 31.7      & 39.2      & 44.3      & 38.4    & 22.4 \\
    \Mymth{-S (8f)}     & R50      & $704\times256$  & 34.2 & 32.6      & 39.9      & 44.7      & 39.1    & 20.7 \\
    \Mymth{-M (8f)}     & R50      & $704\times256$  & 35.6 & 33.7      & 41.1      & 46.0      & 40.3    & 13.4 \\
    \Mymth{-L (8f)}     & R50      & $704\times256$  & 36.2 & 34.7      & 42.1      & 46.7      & 41.2    & 7.2 \\ \midrule
    \bottomrule[0.9pt]
    \end{tabular}}
\end{table}

\textbf{Quantitative Performances.}
In this part, we compare \Mymth{} with previous state-of-the-art methods on the Occ3D-nuScenes dataset.
Our methods not only achieves the superior performances in terms of \riou{} and competitive results in mIoU, but also demonstrates commendable real-time performance.
As depicted in \cref{tab:main}, \Mymth{-T (8f)} reaches 22.4 FPS, significantly faster than dense counterparts and nearly 1.3 times the speed of sparse counterpart SparseOcc (8f).
Despite using only 7 history frames, its $38.4$ \riou{} result easily outperforms other models, including FB-Occ (16f) with \riou{} of $33.5(-4.9)$ and SparseOcc (16f) with \riou{} of $35.1(-3.3)$.
Similarly, \Mymth{-S (8f)} and \Mymth{-M (8f)} achieve a good balance between performance and efficiency.
The heaviest version of \Mymth{} ultimately achieves an \riou{} of 41.2, surpassing the previous best result by a notable margin of 6.1. 

With the same total number of points predicted, we vary the query number and correspondingly change the number of points from each query, leading to different versions of \Mymth{}.
It can be observed that increasing the query number decreases the FPS values from 22.4 to 7.2, while simultaneously boosts model performance in terms of mIoU and \riou{}.
The \Mymth{-M (8f)}, with 2.4K queries, strikes a balance by achieving a comparable RayIoU while maintaining competitive FPS.

Despite the vulnerability of mIoU metric to overestimation manipulations~\cite{liu2023fully},
our \Mymth{} attains a comparable mIoU of 36.2, significantly bridging the gap between dense and sparse models in this metric. These results under different metrics collectively demonstrate the superiority of our \Mymth{}.

\begin{figure}[t]
    \includegraphics[trim={0cm 0cm 0cm 0cm},clip, width=\textwidth]{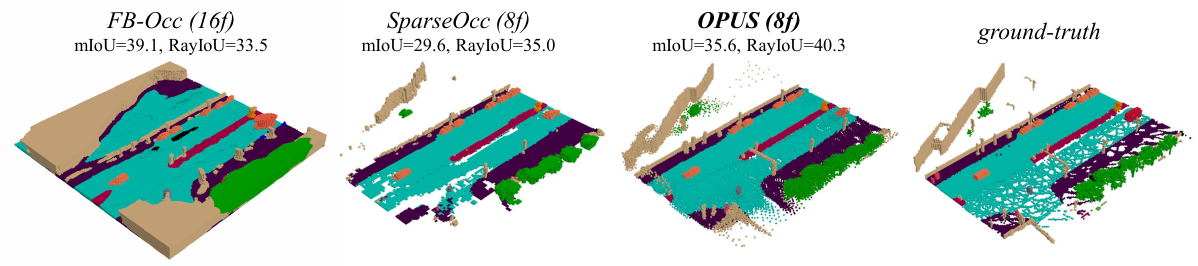}
    \caption{
    Visualizations of occupancy predictions.
    \textbf{Best viewed in color}.
    }\label{fig:main_compare}
     \vspace{-10pt}
\end{figure}

\textbf{Visualization.}
We visualize the predicted occupancy in \cref{fig:main_compare}.
It can be observed that FB-Occ tends to produce denser results compared to sparse methods. 
Though seems complete in the 3D environment, its predicted occupancy results are severely over-estimated, especially for the far areas.
The overestimation may hack the mIoU metric~\cite{liu2023fully}, while heavily penalized by \riou{} that primarily considers the first occupied voxels along rays.
Consequently, FB-Occ achieves the best mIoU of 39.1 but the worst \riou{} value.
On the other hand, SparseOcc occasionally exhibits discontinuous predictions with false negatives, especially in long distances. 
This is attributed to SparseOcc's gradual removal of empty voxels, making erroneous filtering in early stages accumulates and contributes to the final false predictions.
In contrast, our \Mymth{} maintains a more continuous prediction thanks to its end-to-end approach, resulting in a more reasonable visualization.

\subsection{Ablation study and visualizations}\label{sec:exp_ab}

This part details our ablation study and visualizations using the \Mymth{}-M (8f) model.



\def\posrw{CD$_R$}
\def\clsrw{FocalLoss$_R$}
\begin{table}[t]
    \small
    \caption{Model performances with different combinations of proposed strategies.}
    \label{tab:component}
    \centering
    \scalebox{0.9}{
    \begin{tabular}{cccc|c|ccc|>{\columncolor{lightgray!20}}c}
    \toprule[0.9pt]
    \posrw{}   & \clsrw{}   & CPS        & Coarse-to-fine    & mIoU                 & \riou{1m} & \riou{2m} & \riou{4m} & \riou{}              \\ \midrule
               &            &            &                   & 17.4                 & 23.6      & 29.7      & 34.3      & 29.2                 \\ \midrule
    \checkmark &            &            &                   & 23.7 (6.3$\uparrow$) & 23.9      & 30.7      & 35.6      & 30.1 (0.9$\uparrow$) \\
    \checkmark & \checkmark &            &                   & 25.1 (1.4$\uparrow$) & 25.2      & 32.3      & 37.0      & 31.5 (1.4$\uparrow$) \\
    \checkmark & \checkmark & \checkmark &                   & 25.5 (0.4$\uparrow$) & 26.0      & 33.1      & 37.9      & 32.3 (0.8$\uparrow$) \\ 
    \checkmark & \checkmark & \checkmark & \checkmark        & 27.2 (1.7$\uparrow$) & 26.1      & 33.3      & 38.4      & 32.6 (0.3$\uparrow$) \\ \bottomrule[0.9pt]
    \end{tabular}}
\end{table}

\textbf{Effects of the proposed strategies in \Mymth{}.}
In our work, we introduce adaptive re-weighting for the Chamfer distance loss and focal loss, along with consistent point sampling, and coarse-to-fine prediction strategies. 
We examine the impacts of these strategies as shown in \cref{tab:component}.
Without bells and whistles, \Mymth{} achieves a baseline $17.4$ mIoU and a $29.2$ \riou{}.
Replacing the original CD loss into our revision CD$_R$ significantly boosts the mIoU and \riou{} by 6.4 and 0.9, respectively, demonstrating the importance of focusing on erroneous predicted locations in this task.
The FocalLoss$_R$ further improves both metrics by 1.4.
Incorporating the term $\mathbf{\sigma}_p$ in \cref{eq:method_cps} further enhances mIoU and \riou{} by 0.4 and 0.8, demonstrating the efficacy of considering previous point distribution in the current sampling process.
The proposed coarse-to-fine query prediction gradually increases the number of points across the stages.
This scheme not only reduces computations in early stages but also notably benefits model performance, particularly in mIoU, which is increased by 1.7.
These results highlight the cumulative benefits of each component, showcasing how their integration leads to substantial performance gains.


\begin{figure}[t]
    \includegraphics[trim={0cm 0cm 0cm 0cm},clip, width=\textwidth]{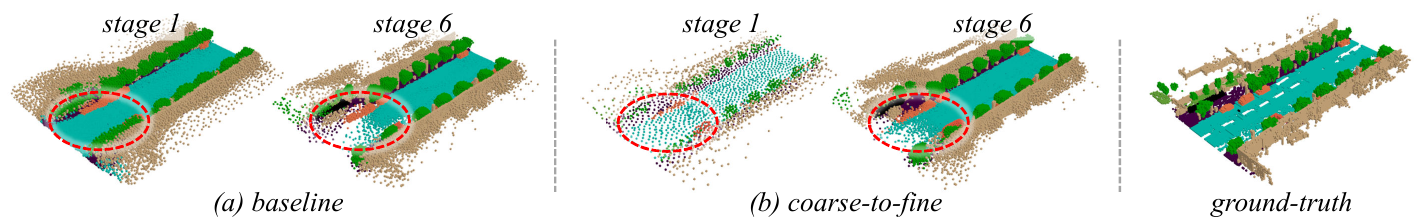}
    \caption{
    Visualizations of the coarse-to-fine predictions.
    }\label{fig:coarsetofine}
     \vspace{-10pt}
\end{figure}

\textbf{Visualization on the coarse-to-fine prediction.}
We visualize the prediction results at different stages in \cref{fig:coarsetofine}.
In the baseline scenario depicted in \cref{fig:coarsetofine}(a), where all decoders regress the same number of points, we observe inconsistent point distributions across stages and numerous false negative predictions in long distances, as highlighted by circles.
This may be attributed to the difficulty of learning the fine-grained occupancy representations in the early stages, impeding the efficient training of the entire framework. 
In contrast, our coarse-to-fine strategy significantly alleviates the learning difficulty in early stages, thereby leading to improved model performances.
As a result,  the point distributions are more consistent among different stages, and the final predictions exhibit much fewer false negatives, as illustrated in \cref{fig:coarsetofine}(b).

\begin{table}[t]
    \small
    \caption{Comparison of various treatments on initial locations $\mathbb{P}^0$.
    "Grid" and "Random" indicate that points are sampled uniformly in BEV space and randomly in the 3D space, respectively
    "Optimized" means that points are randomly initialized but supervised with ground-truths via the CD$_R$ loss.
    }
    \label{tab:initial_pts}
    \centering
    \scalebox{0.9}{
    \begin{tabular}{c|c|ccc|>{\columncolor{lightgray!20}}c}
    \toprule[0.9pt]
    Type       & mIoU & \riou{1m} & \riou{2m} & \riou{4m} & \riou{} \\ \midrule
    Grid       & 22.8 & 22.2      & 28.9      & 33.9      & 28.3 \\
    Random     & 23.1 & 23.6      & 30.5      & 35.6      & 29.9   \\
    Optimized  & 23.7 & 23.9      & 30.7      & 35.6      & 30.1 \\ \bottomrule[0.9pt]
    \end{tabular}}
\end{table}

\begin{wrapfigure}{r}{5cm}
    \centering
    \includegraphics[trim={0cm 0cm 0cm 0cm}, width=0.35\textwidth]{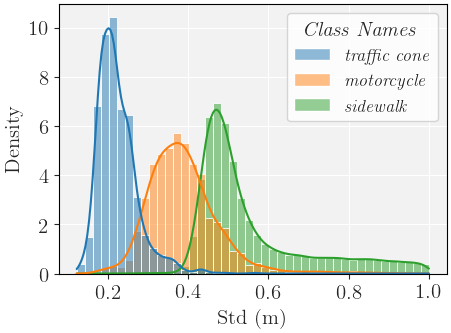}
    \caption{Distributions of standard deviations of points from one query.
    }\label{fig:dist}
     \vspace{-10pt}
\end{wrapfigure}

\begin{figure}[t]
    \includegraphics[trim={0cm 0cm 0cm 0cm},clip, width=\textwidth]{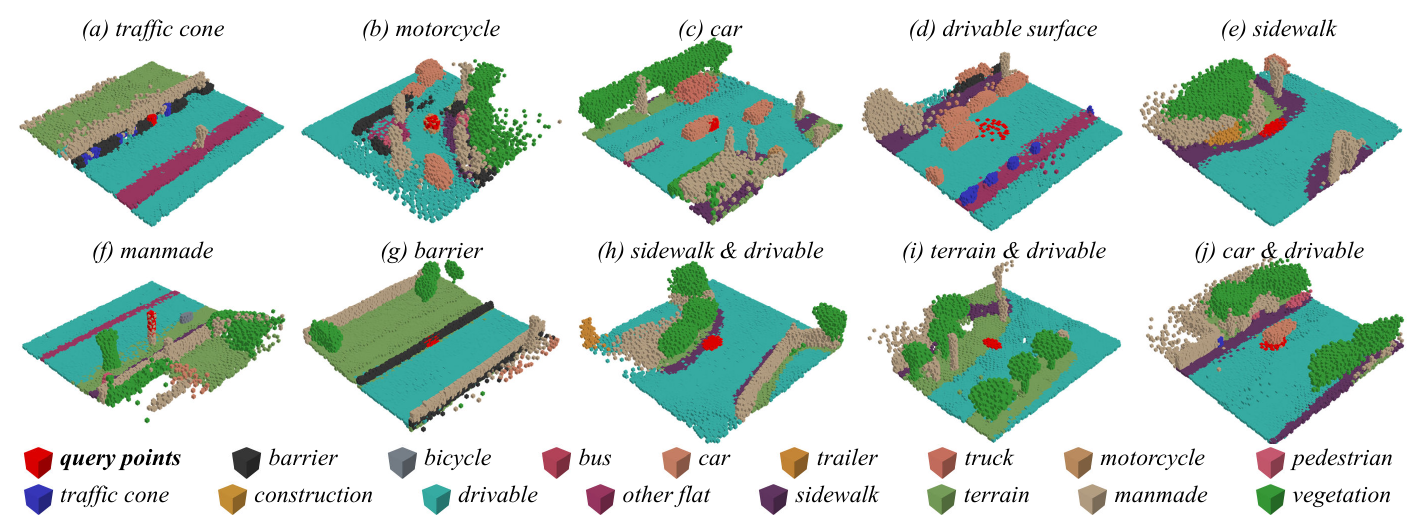}
    \caption{
    Visualizations of points generated from different queries.
    \textbf{Best viewed in color}.
    }\label{fig:pattern}
     \vspace{-10pt}
\end{figure}

\textbf{Visualizations of predicted points.}
In \cref{fig:pattern}, we select a few queries and visualize their predicted points.
Notably, most queries exhibit a tendency to predict points with consistent classes, or even from the same instance, as depicted in \cref{fig:pattern}(a)-(g).
An interesting observation is that the predicted points tend to exhibit diverse distributions in classes with large volumes, such as drivable surfaces and sidewalks.
Conversely, for objects with limited sizes, such as traffic cones, motorcycles, and cars, the points are distributed more closely with respect to the instance size.
The patterns can be further verified by \cref{fig:dist}, where we present the standard deviations of points from queries with three chosen classes.
These results highlight the efficacy of our model in adapting its predictions to the distinct spatial characteristics of various object classes.

As we do not explicitly constrain points from one query to have the same class, it's conceivable that one query could yield points of different classes.
We found this phenomenon commonly occurs at the boundaries between objects.
However, even when classes vary, these points are still closely distributed, as depicted in \cref{fig:pattern}(h)-(j).

\textbf{Influence of treatments on the initial points.}
\cref{tab:initial_pts} compares three different treatments on the initial points $\mathbb{P}^0$.
Grid initialization divides the BEV space into evenly-distributed pillars and orderly assigns pillar centers as the initial locations, a method utilized in BEVFormer~\cite{li2022bevformer}.
Random initialization assigns each location with a uniform distribution in the 3D space.
After initialization, $\mathbb{P}^0$ remains learnable during training.
On top of the random initialization, our \Mymth{} further add supervisions of the ground-truth distributions to $\mathbb{P}^0$ (\ie, $\text{CD}_R(\mathbb{P}^0, \mathbb{P}_g)$ in \cref{eq:final}).
The results in \cref{tab:sparsification} show that random initialization outperforms grid initialization, achieving an mIoU of 23.1 compared to 22.8, and a \riou{} of 29.9 compared to 28.3. 
This improvement is likely due to the fact the random initialization provides a more diverse 3D distribution.
Furthermore, the introduced supervision results in additional improvements of 0.6 on mIoU and 0.2 on \riou{}.
These results reveal the efficiency of the random initialization and the additional supervision on the initial locations.



\begin{figure}[b]
    \includegraphics[trim={0cm 0cm 0cm 0cm},clip, width=\textwidth]{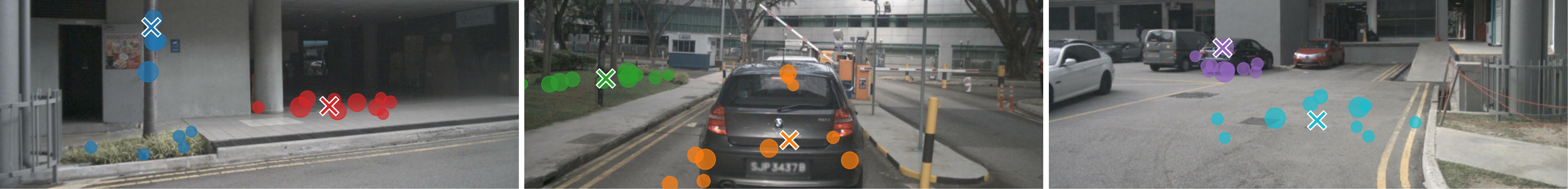}
    \caption{
        the self-attention in decoders.
        for each pivot (marked as $\times$), query points with top 10 attention weights are shown by circles, with sizes proportional to weights.
    \textbf{best viewed in color}.
    }\label{fig:self_attn}
\end{figure}

\textbf{Visualization of the self-attention.}
For better understanding what query points attend to, we visualize top 10 query points with highest self attention weights for each query.
We project them on 2D image for better visualization.
Here are some interesting finding from \cref{fig:self_attn}.
Generally speaking, the query tends to attend to neighbouring query points.
For example, the sidewalk query in the first image and the car query in the second image, allow local information from neighbouring query points flow into, enabling the query to capture detailed local information.
Additionally, the query maintains the ability to attend to semantic related locations even if they are not very close to the query point.
For instance, the vegetation query in the first image of \cref{fig:self_attn} not only attends to the stem of the tree, but also the grass, indicating that the query can capture semantically related information for more accurate predictions.
Another notable observation is that query can sense terrain-related information. 
For instance, in the third image, the sidewalk query points attend to other points along a straight line following the road's edge, highlighting the model's ability to understand scene related structure of the environment.

\textbf{Comparisons between different sparsification strategies}\label{sec:sparsification}
In \cref{tab:sparsification}, we compare \Mymth{} to two other models with different sparsification strategies.
The first baseline is SparseOcc, which achieves sparsification by filtering out empty voxels at various cascade stages.
Following PETRv2~\cite{liu2023petrv2}, the second baseline is a pillar-patch based method that partitions the 3D space into a small number of pillar-patches.
We use $50 \times 50$ queries with each corresponding to the classification of neighbouring $4 \times 4 \times 16$ voxels.
For a fair comparison, all these models are trained for 100 epochs.
In contrast, our model achieve best results after sufficient training with a \riou{} score of 38.0,far outperforming SparseOcc with a \riou{} score of 34.3.
On the other hand, our model can also runs in a real-time speed.
These results demonstrates the superiority of our sparification procedure.

\textbf{Comparisons on the Waymo-Occ3D dataset.} 
We further simply implement \Mymth{} on the Waymo-Occ3D~\cite{sun2020scalability} dataset to explore the generalization and robustness of \Mymth{}.
As Waymo-Occ3D is not commonly used as a standard benchmark for vision-centric approaches, the only vision-based method we found with reported results on this dataset is the Occ3D paper, which evaluates BEVDet, TPVFormer, BEVFormer, and the newly proposed CTF-Occ~\cite{tian2024occ3d}.
We trained the \Mymth{-L} (1f) on 20\% of the dataset for a fair comparison with these baselines.
As reported in \cref{tab:waymo}, despite not fine-tuning the training configurations, \Mymth{-L} already achieves 19.0 mIoU, outperforming all previous methods.
Moreover, \Mymth{-L} also reaches 8.5 FPS on the Waymo-Occ3D dataset, which is around 3 times the speed of CTF-Occ and 2 times the speed of BEVFormer.

\begin{table}[t]
    \small
    \caption{Comparisons between different sparsification strategies.}
    \label{tab:sparsification}
    \centering
    \scalebox{0.83}{
    \begin{tabular}{c|cc|ccc|c|c}
    \toprule[0.9pt]
    Model                     & $Q$   & $R$                               & \riou{1m} & \riou{2m} & \riou{4m} & \riou{} & FPS \\ \midrule
    SparseOcc                 & \multicolumn{2}{c|}{(4000/16000/64000)}   & 28.4      & 34.9      & 39.6      & 34.3    & 17.3\\ 
    PETR v2                   & 2500  & 256                               & 24.4      & 31.0      & 36.3      & 30.6    & 13.8 \\ 
    \Mymth{}                  & 2400  & 32                                & 31.7      & 38.8      & 43.4      & 38.0    & 13.4 \\ \bottomrule[0.9pt]
    \end{tabular}}
    \vspace{-5pt}
\end{table}

\begin{table}[t]
    \small
    \caption{Performance on the Waymo-Occ3D dataset.}
    \label{tab:waymo}
    \centering
    \setlength{\tabcolsep}{2.4pt}
    \scalebox{0.83}{
    \begin{tabular}{c|ccccccccccccccc|>{\columncolor{lightgray!10}}c>{\columncolor{lightgray!10}}c>{\columncolor{lightgray!10}}c}
    \toprule[0.9pt]
    Model      & \V{General} & \V{Vehicle} & \V{Bicyclist} & \V{Ped.} & \V{Sign} & \V{Tfc. light} & \V{Pole} & \V{Cons. cone} & \V{Bicycle} & \V{Motorcycle} & \V{Building} & \V{Vegetaion} & \V{Treetrunk} & \V{Road} & \V{Sidewalk} & \V{mIoU} & \V{\riou{}} & \V{FPS} \\ \midrule
    BEVDet     & 0.13 & 13.06 & 2.17  & 10.15 & 7.80  & 5.85  & 4.62  & 0.94  & 1.49  & 0.0  & 7.27  & 10.06 & 2.35 & 48.15 & 34.12 & 9.88  & -    & -   \\
    TPVFormer  & 3.89 & 17.86 & 12.03 & 5.67  & 13.64 & 8.49  & 8.90  & 9.95  & 14.79 & 0.32 & 13.82 & 11.44 & 5.8  & 73.3  & 51.49 & 16.76 & -    & -   \\
    BEVFormer  & 3.48 & 17.18 & 13.87 & 5.9   & 13.84 & 2.7   & 9.82  & 12.2  & 13.99 & 0.0  & 13.38 & 11.66 & 6.73 & 74.97 & 51.61 & 16.76 & -    & 4.6 \\ 
    CTF-Occ    & 6.26 & 28.09 & 14.66 & 8.22  & 15.44 & 10.53 & 11.78 & 13.62 & 16.45 & 0.65 & 18.63 & 17.3  & 8.29 & 67.99 & 42.98 & 18.73 & -    & 2.6 \\ 
    \Mymth{-L} & 4.66 & 27.07 & 19.39 & 6.53  & 18.66 & 6.41  & 11.44 & 10.40 & 12.90 & 0.0  & 18.73 & 18.11 & 7.46 & 72.86 & 50.31 & 19.00 & 24.7 & 8.5 \\ \bottomrule[0.9pt]
    \end{tabular}}
    \vspace{-5pt}
\end{table}

\section{Conclusions and limitations}

This paper introduces a novel perspective on occupancy prediction by framing it as a direct set prediction problem. 
Using a transformer encoder-decoder architecture, the proposed \Mymth{} directly predicts occupied locations and classes in parallel from a set of learnable queries. 
The matching between predictions and ground truths is accomplished through two efficient tasks in parallel, facilitating end-to-end training with a large number of points in this application.
In addition, the query features are enhanced via a list of non-trivial designs (\ie, coarse-to-fine learning, consistent point sampling, and loss re-weighting), and therefore leads to boosted prediction performances.
Our experiments on the Occ3D-nuScenes benchmark demonstrate that \Mymth{} surpasses all prior arts in terms of both accuracy and efficiency, thanks to the sparse designs in our framework.

However, the proposed \Mymth{} also comes with new challenges, particularly regarding the convergence speed. 
The slow convergence may potentially be alleviated by drawing lessons from follow-up works of DETR, which have largely addressed the convergence issue of the original DETR.
Another challenge is that while sparse approaches typically achieve higher \riou{} compared to dense counterparts, they often struggle with the mIoU metric. 
Improving the mIoU performance while maintaining superior RayIoU results is a promising direction for future works.
Moreover, despite conducting experiments on vision-only datasets, our core formulation is directly applicable to multi-modal tasks as well. We leave the multi-modal occupancy prediction as future work.



\clearpage

\section*{Acknowledgments}
This research was supported by NSFC (NO. 62225604, NO. 62276145), the Fundamental Research Funds for the Central Universities (Nankai University, 070-63223049) and partially supported by NSFC (No. 62376153, 62402318, 24Z990200676). Computations were supported by the Supercomputing Center of Nankai University (NKSC).

{
\small
\bibliographystyle{plainnat}
\bibliography{egbib}
}

\newpage
\appendix




\section*{Appendix}

\section{Broader impacts}\label{sec:impact}
Our work proposes an end-to-end paradigm for occupancy prediction, achieving state-of-the-art RayIoU performance with fast inference speeds. 
This advancement can lead to real-time and precise occupancy outcomes, which are crucial for real-world applications of autonomous driving (AD). 
Consequently, the most significant positive impact of our work is the enhancement of safety and response speed in AD systems.

However, the biggest negative societal impact of this work, as with any component of AD systems, is the safety concern. 
Autonomous driving systems are directly related to human lives, and erroneous predictions or decisions can lead to hazardous outcomes.
Therefore, increasing the accuracy of occupancy outcomes and developing complementary methods to address false predictions will require substantial follow-up efforts.

\section{Licenses for involved assets}\label{sec:license}

Our code is built on top of the codebase\footnote[1]{https://github.com/MCG-NJU/SparseBEV} provided by SparseBEV~\cite{liu2023sparsebev}, which is subject to the MIT license.
Our experiments are conducted on the Occ3D-nuScenes~\cite{tian2024occ3d} which provides occupancy labels for the nuScenes dataset~\cite{caesar2020nuscenes}.
Occ3D-nuScenes is licensed under the MIT license, and nuScenes is licensed under the CC BY-NC-SA 4.0 license.

\section{Complexity analysis}\label{sec:complex}
In this part, we provide a detailed analysis of the time and space complexity involved in matching $m$ predictions with $n$ ground-truths.

\textbf{Hungarian algorithm.}
The Hungarian algorithm's core involves finding augmenting paths for $\min(m, n)$ iterations.
Each iteration can be visualized as an attempt to improve the current matching by finding the shortest augmenting path in the residual graph, which has a complexity of $O(\max(m, n)^2)$ using with Dijkstra's algorithm.
Consequently, the time complexity for the Hungarian algorithm is $O(\min(m, n)\cdot \max(m,n)^2)$.

Meanwhile, the Hungarian Algorithm necessitates computing a cost matrix of size $m\times n$ to store the costs linked with each potential assignment.
Throughout the matching process, the tracked labels and matched pairs each demand $O(\min(m, n))$ space. Hence, the final space complexity is $O(m\times n)$.

\textbf{Our method.}
Our method employs the Chamfer distance loss, which involves computing pair-wise distances and determining the smallest distance for each point. 
The first step requires a time complexity of $O(m\times n)$ and the next step requires $O(m\times n)$ as well.
The assignment of semantic labels can re-use the results of previous nearest search, therefore requires no additional computations.
In the end, the time complexity is $O(m\times n)$.

For each point in one set, the algorithm needs to keep track of the minimum distance to any point in the other set. 
This can be done using a single variable per point, resulting in $O(m)$ and $O(n)$ in the respective directions.
Semantic label assignment, meanwhile, incurs a space complexity of $O(m)$. 
Collectively, this sums up to $O(2m+n)$.

\textbf{Comparison of the two methods.}
In conclusion, when $m$ and $n$ are comparable in scale, the Hungarian algorithm exhibits time complexity of $O(n^3)$ and space complexity of $O(n^2)$, whereas our method demonstrates significantly improved efficiencies with complexities of $O(n^2)$ and $O(n)$, respectively. 
This represents a notable reduction in both time and space requirements, making it a more efficient solution for large-scale applications.


\section{Additional experiments.}\label{sec:supp_exp}

\subsection{Comparison of Hungarian matching and our method}

\begin{table}[h]
    \caption{Comparison of Hungarian algorithm and our label assignment scheme.}
    \label{tab:comp_matching}
    \small
    \centering
    \scalebox{1.0}{
    \begin{tabular}{c|cc|cc}
    \toprule[0.9pt]
    Number  & \multicolumn{2}{c|}{Time (ms)} & \multicolumn{2}{c}{GPU (Mb)} \\
    of Points& Hungarian Algorithm & Ours & Hungarian  Algorithm & Ours \\
    \midrule 
    100   & 0.52 & 0.12 & 39 & 14\\
    1,000 & 78.34 & 0.13 & 81 & 14 \\
    10,000 & 24,216.35 & 1.25 & 2,304 & 15\\
    100,000 & - & 28.85 & - & 39\\
    \bottomrule
    \end{tabular}
    }
\end{table}

\cref{tab:comp_matching} presents a comparison of the duration and GPU utilization when matching two point clouds with the same number of points. 
It is evident that the Hungarian algorithm exhibits scalability issues. 
For instance, when the point number is 10K, it consumes approximately 24 seconds and 2,304Mb of GPU memory for a single matching. 
Scaling up to 100K points renders the matching infeasible due to CUDA memory constraints, even on an 80G A100 GPU.

In contrast, our label assignment method achieves remarkable efficiency, requiring only about 1.25ms and 28.85ms for 10K and 100K points, respectively. 
Furthermore, the GPU memory consumption during training is negligible. 
These findings reveal the practicality and efficacy of our label assignment approach, particularly for the occupancy prediction where point counts can easily exceed 10K.

\begin{table}[ht]
    \small
    \caption{Configurations for different models.}
    \label{tab:setting}
    \centering
    \scalebox{0.9}{
    \begin{tabular}{c|cc|cccccc}
    \toprule[0.9pt]
    \multirow{2}{*}{Model} & \multirow{2}{*}{$Q$}  & \multirow{2}{*}{$S$} & \multicolumn{6}{c}{point number}   \\
                           &                       &                      & s1 & s2 & s3 & s4 & s5 & s6  \\ \midrule
    \Mymth{}-T             & 600                   & 4                    & 1  & 4  & 16 & 32 & 64 & 128 \\
    \Mymth{}-S             & 1200                  & 2                    & 1  & 4  & 8  & 16 & 32 & 64  \\
    \Mymth{}-M             & 2400                  & 2                    & 1  & 2  & 4  & 8  & 16 & 32  \\
    \Mymth{}-L             & 4800                  & 2                    & 1  & 2  & 4  & 8  & 16 & 16  \\ \bottomrule[0.9pt]
    \end{tabular}}
\end{table}

\subsection{Detailed configuration for different versions.}\label{sec:config}
In this section, we detail the settings of various versions of our model, as shown in \cref{tab:setting}, each tailored to prioritize different aspects of performance and speed.
Our fastest model \Mymth{}-T utilizes only 0.6K queries, with each query sampling 4 points in images.
The number of predicted points are  1, 4, 16, 32, 64 and 128 for 6 stages, respectively. 
This configuration ensures a rapid processing time while maintaining competitive performance.
Other versions of our model, such as \Mymth{}-S, \Mymth{}-M, \Mymth{}-L, sample 2 points in CPS module, progressively double the number of queries and adjust the number of predicted points accordingly to balance speed and accuracy.
All these models predict the same amount of points in the end. 

\begin{table}[ht]
    \small
    \caption{Performance with different points predicted.}
    \label{tab:refine_pts}
    \centering
    \scalebox{0.9}{
    \begin{tabular}{c|c|c|ccc|>{\columncolor{lightgray!10}}c}
    \toprule[0.9pt]
    Model                       & point number & mIoU & \riou{1m} & \riou{2m} & \riou{4m} & \riou{} \\ \midrule
    \multirow{4}{*}{\Mymth{}-M} & 64           & 28.4 & 22.2      & 29.5      & 34.8      & 28.8 \\
                                & 32           & 27.2 & 26.1      & 33.3      & 38.4      & 32.6 \\
                                & 16           & 22.8 & 28.1      & 35.3      & 40.2      & 34.5 \\ 
                                & 8            & 16.4 & 27.4      & 34.6      & 39.6      & 33.9 \\ \bottomrule[0.9pt]
    \end{tabular}}
\end{table}

\subsection{Effects of various refined points number in last layer.}\label{sec:number_p}
\cref{tab:refine_pts} assesses the impact of varying the number of predicted points in the last layer.
We use \Mymth{}-M as our model for this experiment.
As shown in the table, mIoU steadily rises as the number of points increase from 8 to 64, going from 16.4 to 28.4. 
This trend is expected since increasing the number of points generally leads to higher mIoU by covering more voxels, as mIoU penalizes false negative (FN) heavily.
However, the \riou{} results peak when model predicting 16 points and decline with further increasing points. 
This decline occurs partly because adding more points beyond a certain extent introduces noise, which negatively impacts \riou{}, which emphasizes first occupied voxels along the ray.

\begin{table}[ht]
    \small
    \caption{Performance across different distances.}
    \label{tab:distance}
    \centering
    \scalebox{0.9}{
    \begin{tabular}{c|cccc}
    \toprule[0.9pt]
    Model      & overall & $0m \sim 20m$ & $20m \sim 40m$ & $>40m$  \\ \midrule
    FB-Occ     & 33.5    & 41.3          & 24.2           & 12.1    \\
    \Mymth{-L} & 41.2    & 49.10         & 31.15          & 13.73   \\ \bottomrule[0.9pt]
    \end{tabular}}
\end{table}

\subsection{Predictions across different distances}
We report the RayIoU of FB-Occ and \Mymth{} at different ranges in \cref{tab:distance}. It is evident that \Mymth{} demonstrates a more pronounced advantage in nearby areas than at far distances. This could be  attributed to the phenomenon pointed out by SparseOcc: dense approaches tend to overestimate the surfaces, especially in nearby areas.

\section{Additional qualitative analysis}

\subsection{Differences between SparseOcc and \Mymth{}}

\textbf{View perspective of occupancy prediction.}
The fundamental difference lies in the perspective of occupancy prediction.
As depicted in the main draft, all previous methods, including SparseOcc~\cite{liu2023fully}, treat occupancy prediction as a standard classification task.
\Mymth{}, however, pioneers a set prediction viewpoint, offering a novel, elegant, and end-to-end sparsification approach.

\textbf{Multi-stage vs. end-to-end sparsification procedure.}
SparseOcc generates sparse occupancy by gradually discarding voxels through multiple stages.
The discarding of empty voxels at early stages is irreversible, leading to obvious cumulative errors, as illustrated in \cref{fig:main_compare}. 
Conversely, \Mymth{} circumvents complex filtering mechanisms by directly predicting a sparse set, resulting in more coherent outcomes.

\textbf{Detailed model design.} In terms of a more detailed perspective of the structure, there are also many differences such as:
\begin{itemize}
    \item \textbf{Query number.} In NuScene-Occ3D, SparseOcc necessitates 32K queries in its final stage. \Mymth{}, by comparison, operates with a mere 0.6K-4.8K queries for occupancy prediction, capitalizing on its flexible nature and contributing to its fast inference pace.
    \item \textbf{Coarse-to-fine procedure.} SparseOcc's coarse-to-fine strategy involves progressively filtering empty voxels and subdividing occupied voxels into finer ones. In contrast, \Mymth{} interprets coarse-to-fine as the escalation in number of predicted points across stages.
    \item \textbf{Learning objective.} Our learning target encompasses predicting both semantic classes and occupied locations, simultaneously. The latter is a new objective introduced by \Mymth{}, achieved through a modified Chamfer distance loss.
\end{itemize}

\begin{figure*}[th]
    \centering
    \includegraphics[trim={0cm 0cm 0cm 0cm},clip, width=0.9\textwidth]{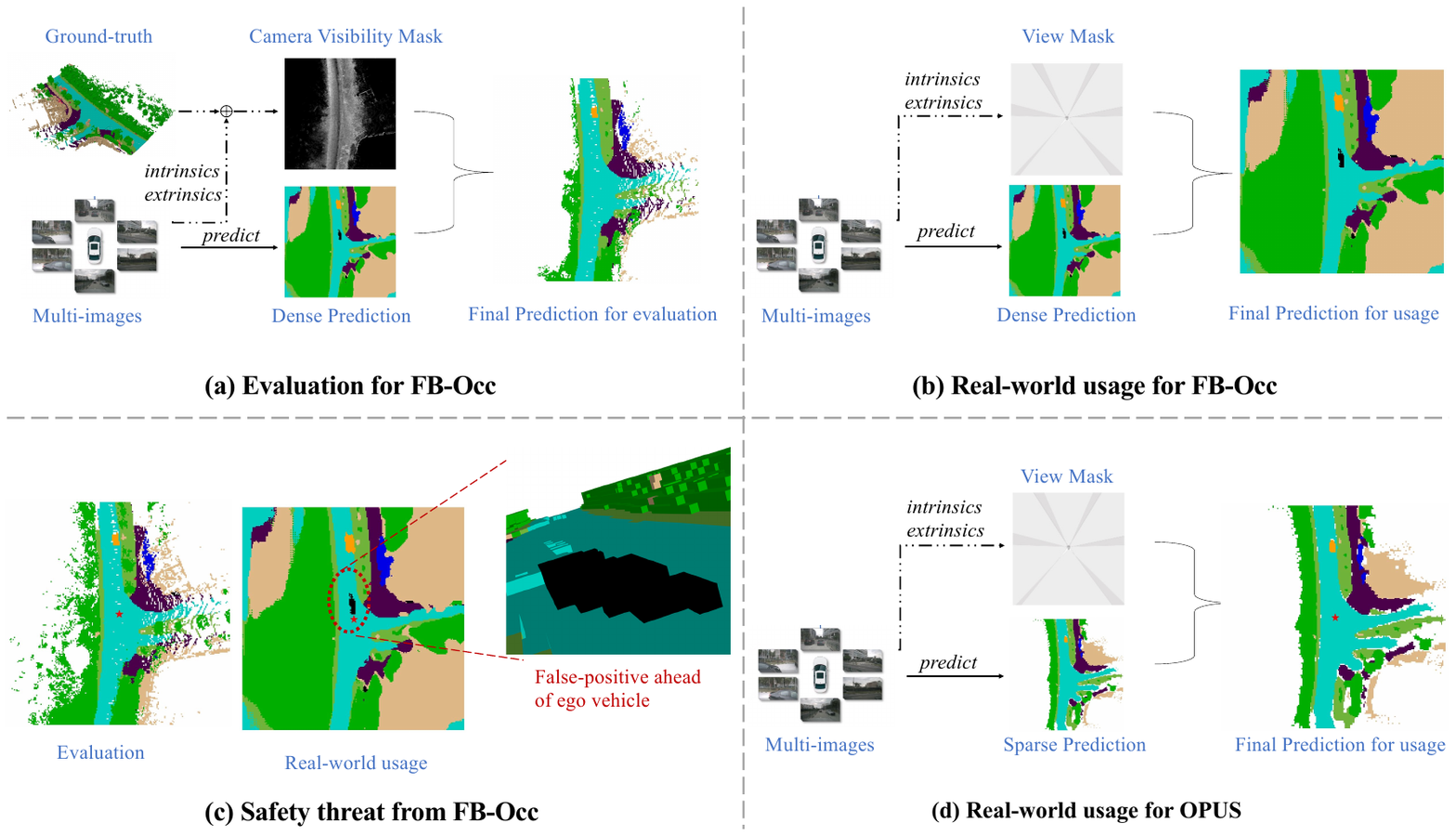}
    \vspace{-15pt}
    \caption{
    Illustration of safety threat due to discrepancies between evaluation metrics and real-world scenarios. 
    (a) Before evaluation, the camera visibility mask is first generated according to camera intrinsics and extrinsics.
    Then, the dense prediction will be masked to get the final prediction for evaluation.
    (b) For real-world usage, we cannot have camera visibility reasoning without knowing the ground-truth occupancy.
    We can only generate the view mask from camera intrinsics and extrinsics, which fails to filter out the over-estimated voxels from dense models.
    (c) Plenty of false positive predictions are made close to the ego vehicle, marked by the symbol of red star.
    These erroneously predicted voxels are filtered during evaluating mIoU, but could cause hazardous safety issue.
    (d) The \Mymth{} produces sparse occupancy predictions and suffers much less from the over-estimation.
    Consequently, no such safety threat occurs in this scenario. 
    \textbf{Best viewed in color}.
    }\label{fig:usage}
    \vspace{-10pt}
\end{figure*}

\begin{figure*}[th]
    \centering
    \includegraphics[trim={0cm 0cm 0cm 0cm},clip, width=0.9\textwidth]{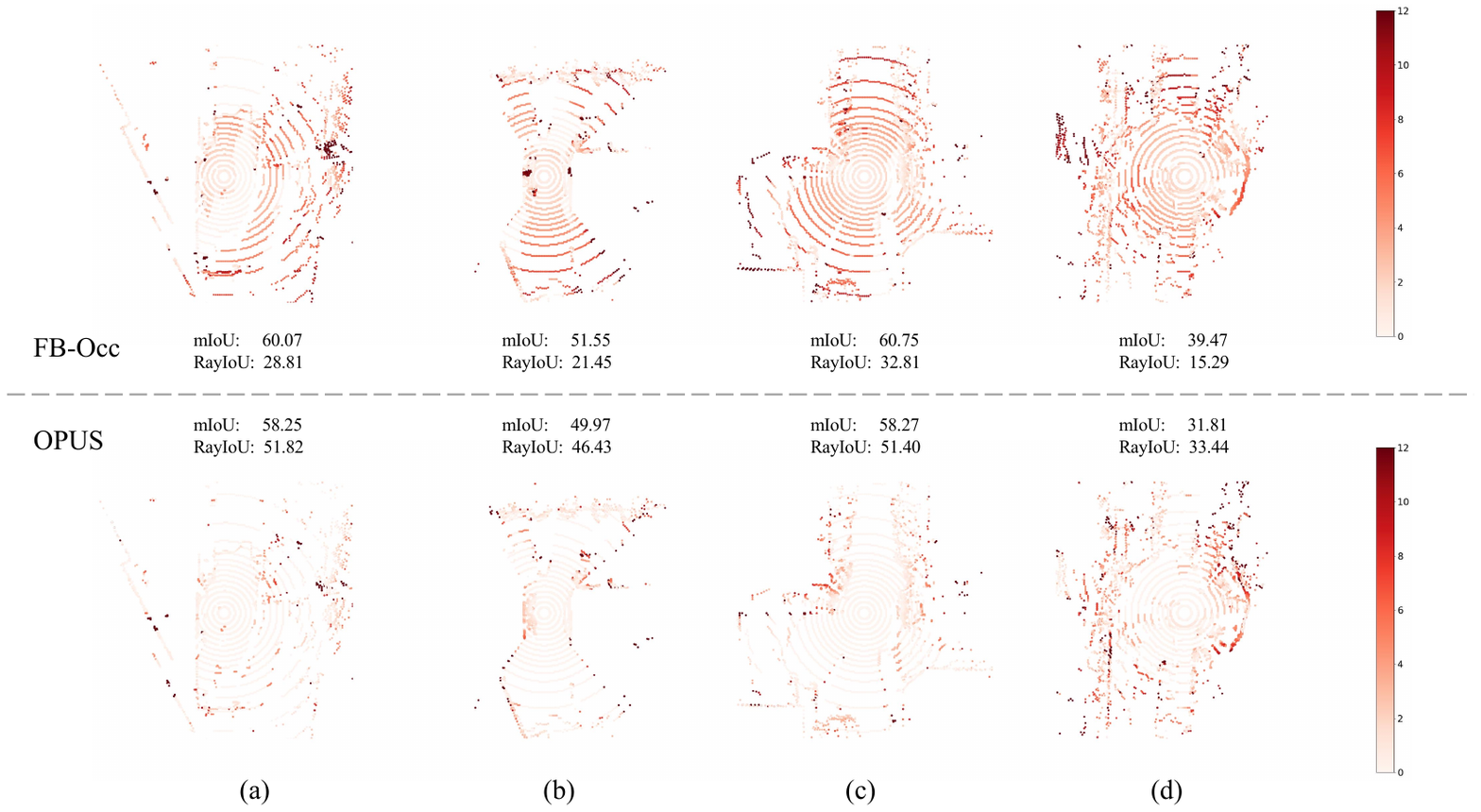}
    \caption{
    The predicted error maps of FB-Occ and \Mymth{}.
    When compared with FB-Occ, \Mymth{} has lower mIoU and higher RayIoU results, and achieves evidently smaller errors.
    \textbf{Best viewed in color}.
    }\label{fig:error_map}
\end{figure*}

\subsection{Analysis of relationships between mIoU, RayIoU and driving safety.}

Our \Mymth{-L (8f)} has achieved a state-of-the-art RayIoU of 41.17, outperforming the previous sparse model SparseOcc by 6.07 and the dense model FB-Occ by 7.7.
The mIoU gap between sparse and dense methods is also reduced from 8.5 in SparseOcc to 3.0 in \Mymth{}.
However, the implications of this gap on safety remain ambiguous.
This concern is particularly pertinent in the context of autonomous driving, and we would like to clarify this as follows:

\textbf{Risks of dense predictions.}
The biggest issue of dense predictions is the discrepancies between evaluation metrics and real-world scenarios.
As shown in \cref{fig:usage}, evaluation metrics only consider voxels within the camera mask, which is derived from camera parameters and ground truth.
However, in real-world applications, we can only produce view mask based on camera intrinsics and extrinsics, failing to filtering out over-estimated voxels.
From \cref{fig:usage} and and \cref{fig:main_compare}, dense methods can misidentify occupied voxels, even close to the ego vehicle.
These errors are overlooked during evaluation but pose significant safety hazards in real-world scenarios. In contrast, \Mymth{} suffer much less from this issue as it does not over-estimate occupancy.

\textbf{The depth errors of \Mymth{} is much smaller than FB-Occ.}
In \cref{fig:error_map}, we compare the depth errors of FB-Occ and \Mymth{} along camera rays.
\Mymth{} demonstrates lower depth errors across all scenes, despite its relatively low mIoU performance.
Given the significance of the first occupied voxel for safety, \Mymth{}'s precision in this regard enhances safety rather than detracting from it.

In conclusion, while it is necessary to minimize the mIoU gap between sparse and dense methods, our analysis indicates that mIoU might not fully represent potentially hazardous situations. Therefore, it would be more rational to take both mIoU and RayIoU into consideration for the occupancy task.

\clearpage

\subsection{Occupancy predictions of different methods.}
\begin{figure*}[h]
    \centering

    \includegraphics[trim={0pt 0pt 0pt 0pt},clip, width=0.24\textwidth]{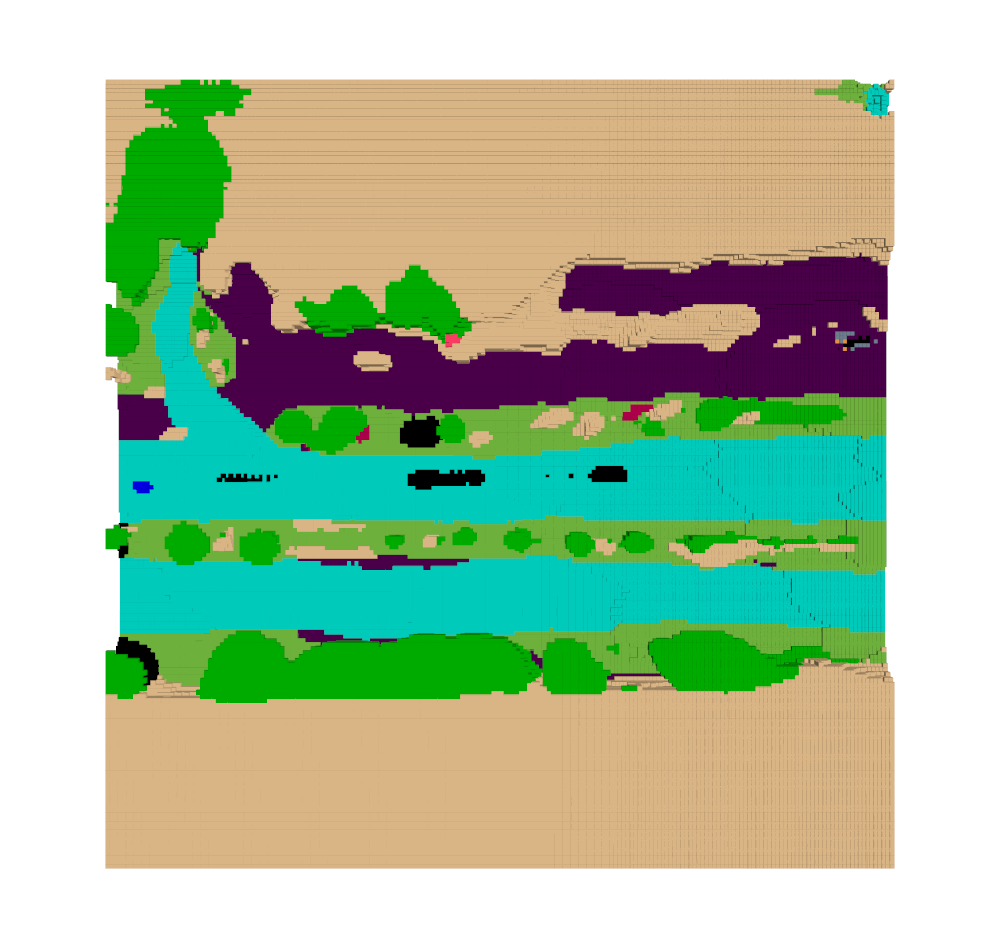}
    \includegraphics[trim={0pt 0pt 0pt 0pt},clip, width=0.24\textwidth]{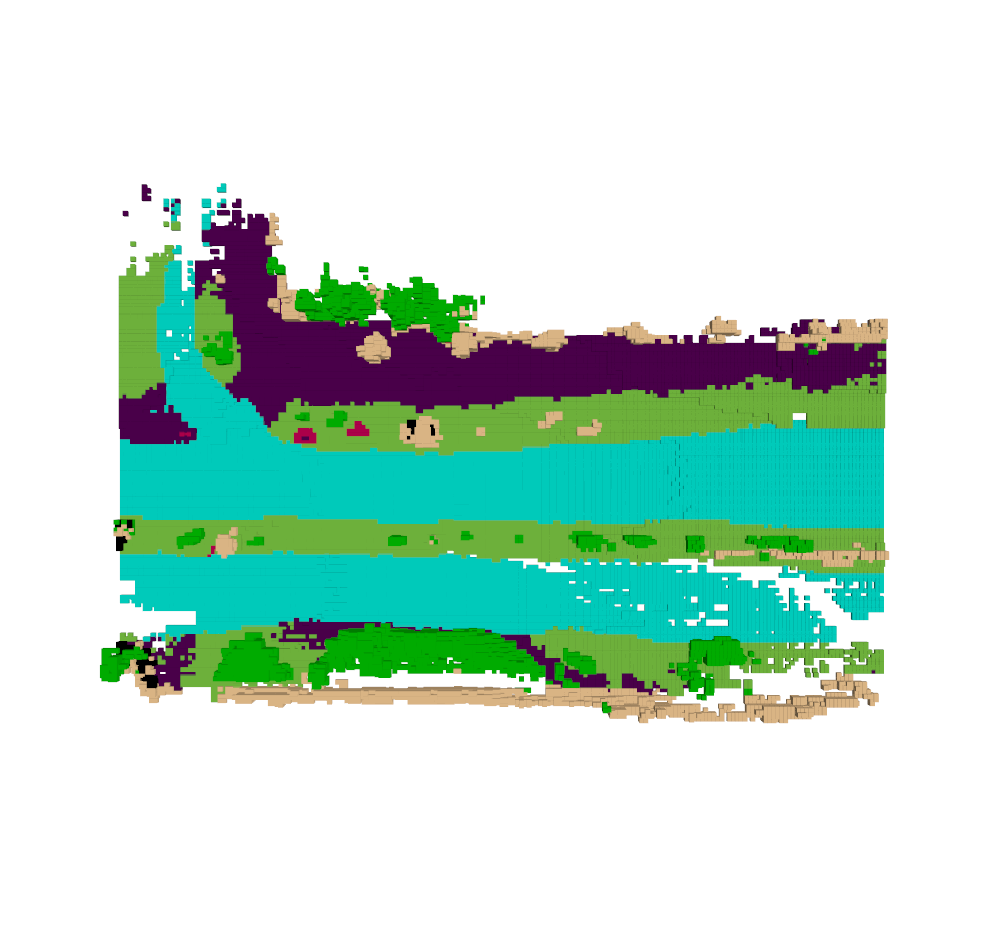}
    \includegraphics[trim={0pt 0pt 0pt 0pt},clip, width=0.24\textwidth]{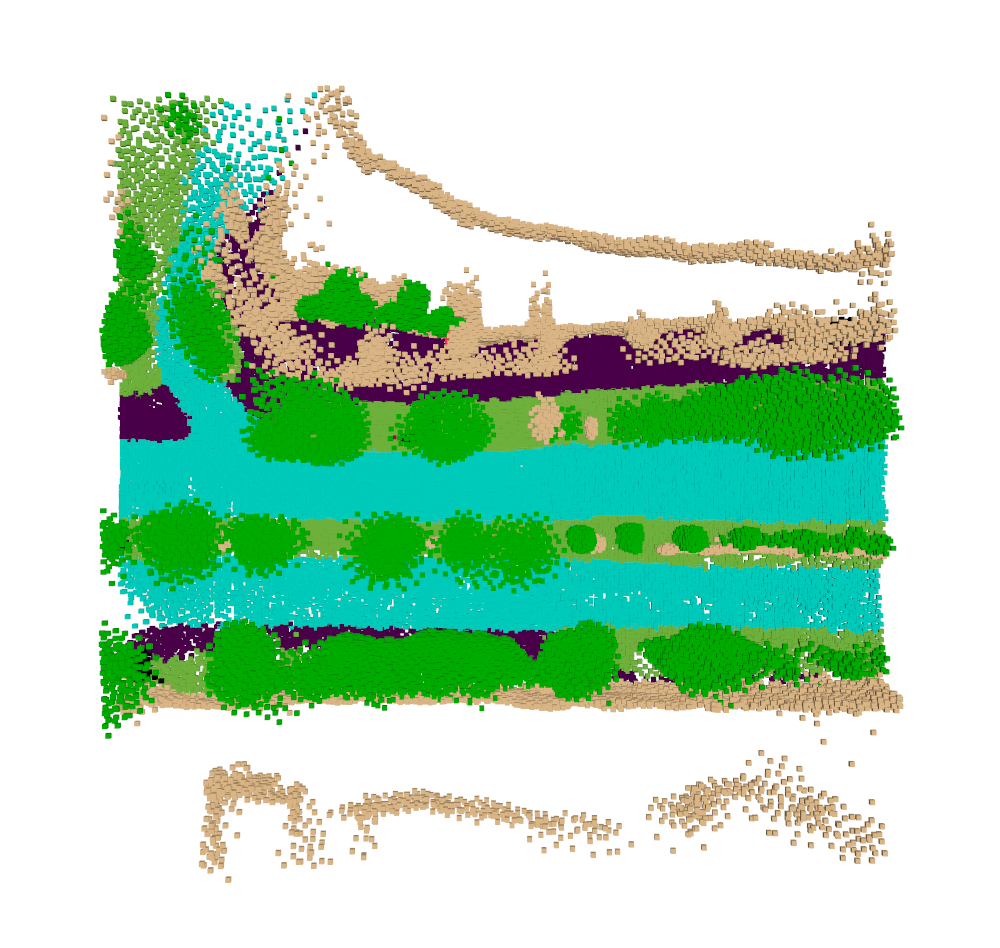}
    \includegraphics[trim={0pt 0pt 0pt 0pt},clip, width=0.24\textwidth]{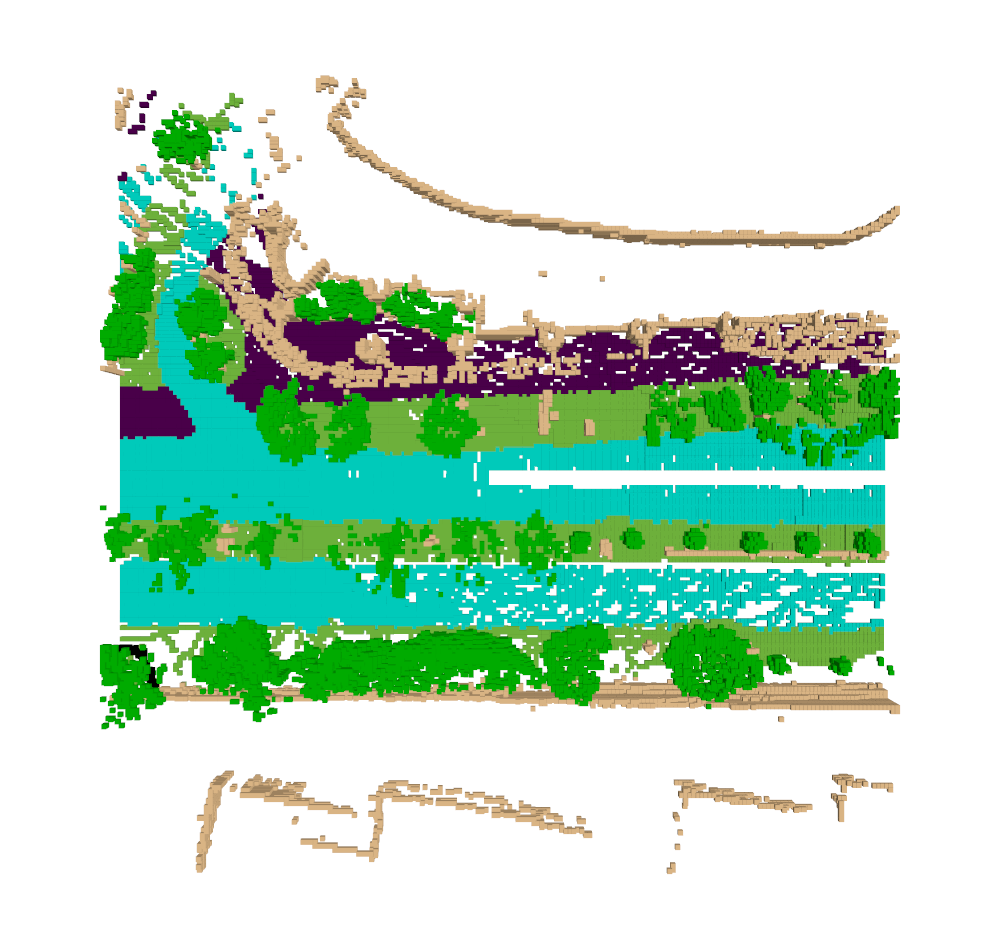}
    \put(-363, 90){\footnotesize (a) FB-Occ}
    \put(-268, 90){\footnotesize (b) SparseOcc}
    \put(-160, 90){\footnotesize (c) \Mymth{}}
    \put(-80, 90){\footnotesize (d) ground-truth}

    \includegraphics[trim={0pt 0pt 0pt 0pt},clip, width=0.24\textwidth]{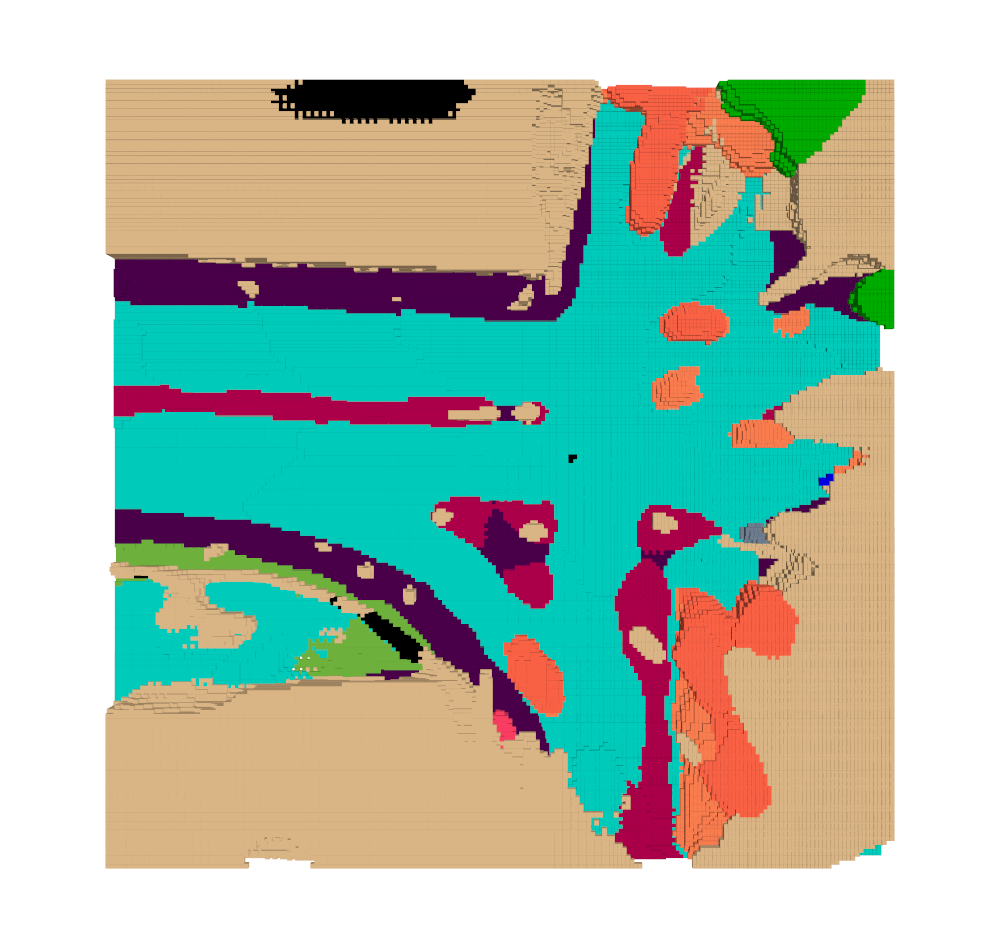}
    \includegraphics[trim={0pt 0pt 0pt 0pt},clip, width=0.24\textwidth]{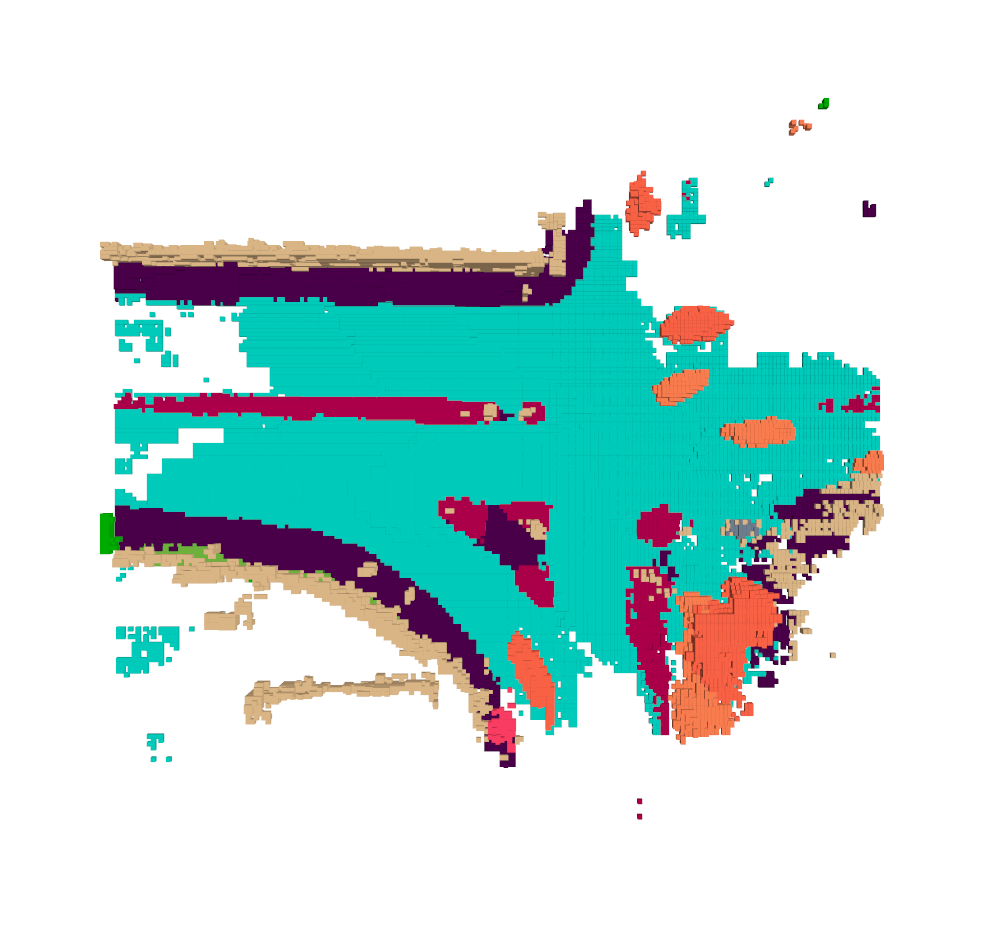}
    \includegraphics[trim={0pt 0pt 0pt 0pt},clip, width=0.24\textwidth]{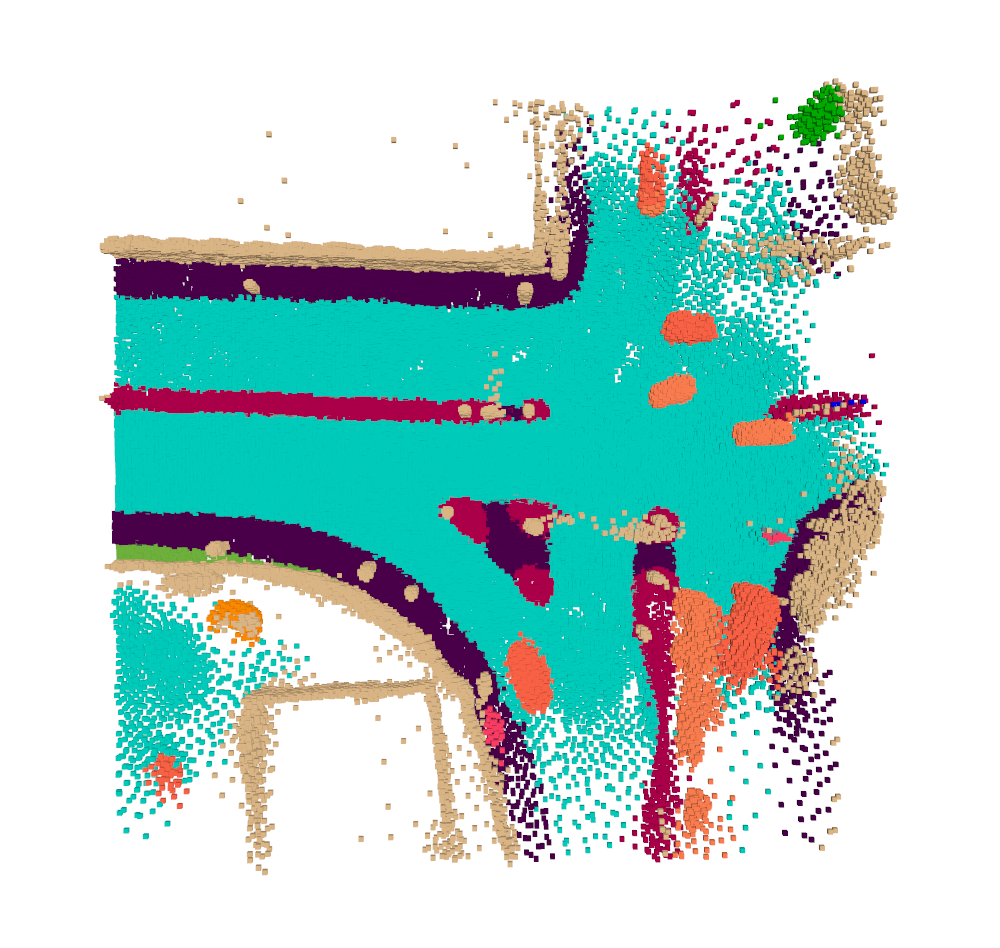}
    \includegraphics[trim={0pt 0pt 0pt 0pt},clip, width=0.24\textwidth]{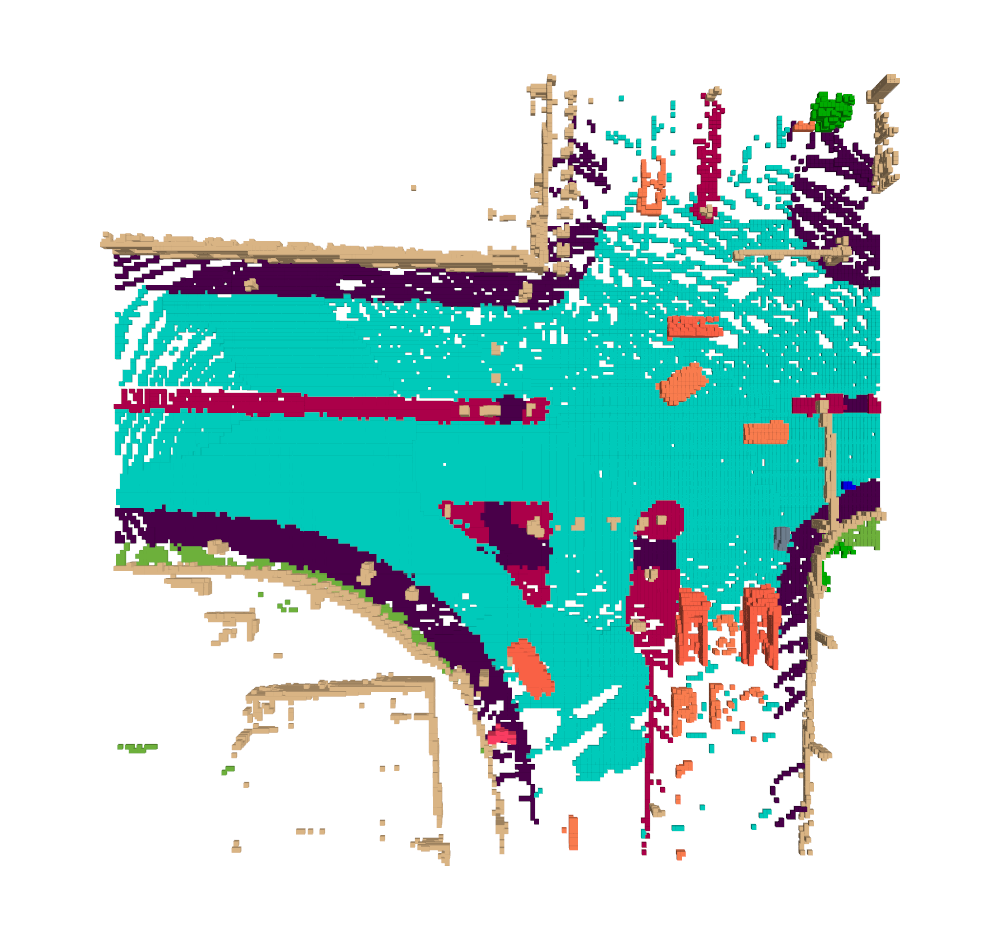}
    
    \includegraphics[trim={0pt 0pt 0pt 0pt},clip, width=0.24\textwidth]{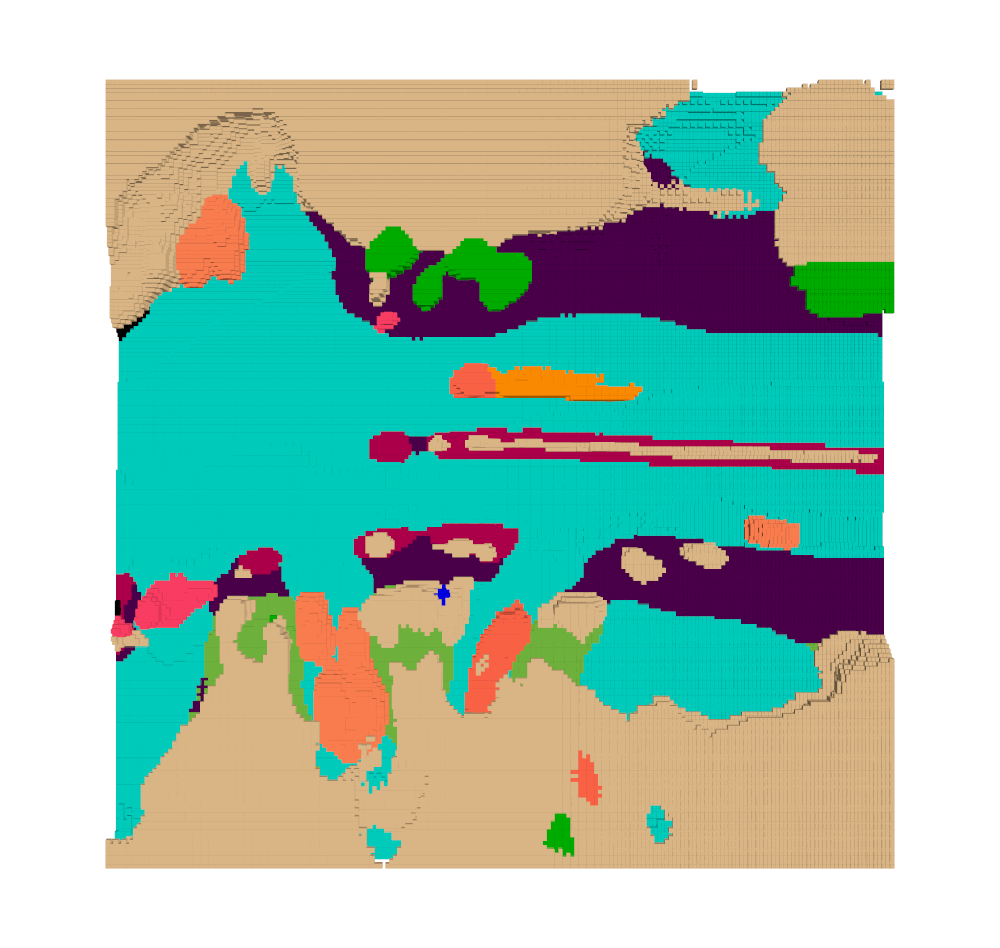}
    \includegraphics[trim={0pt 0pt 0pt 0pt},clip, width=0.24\textwidth]{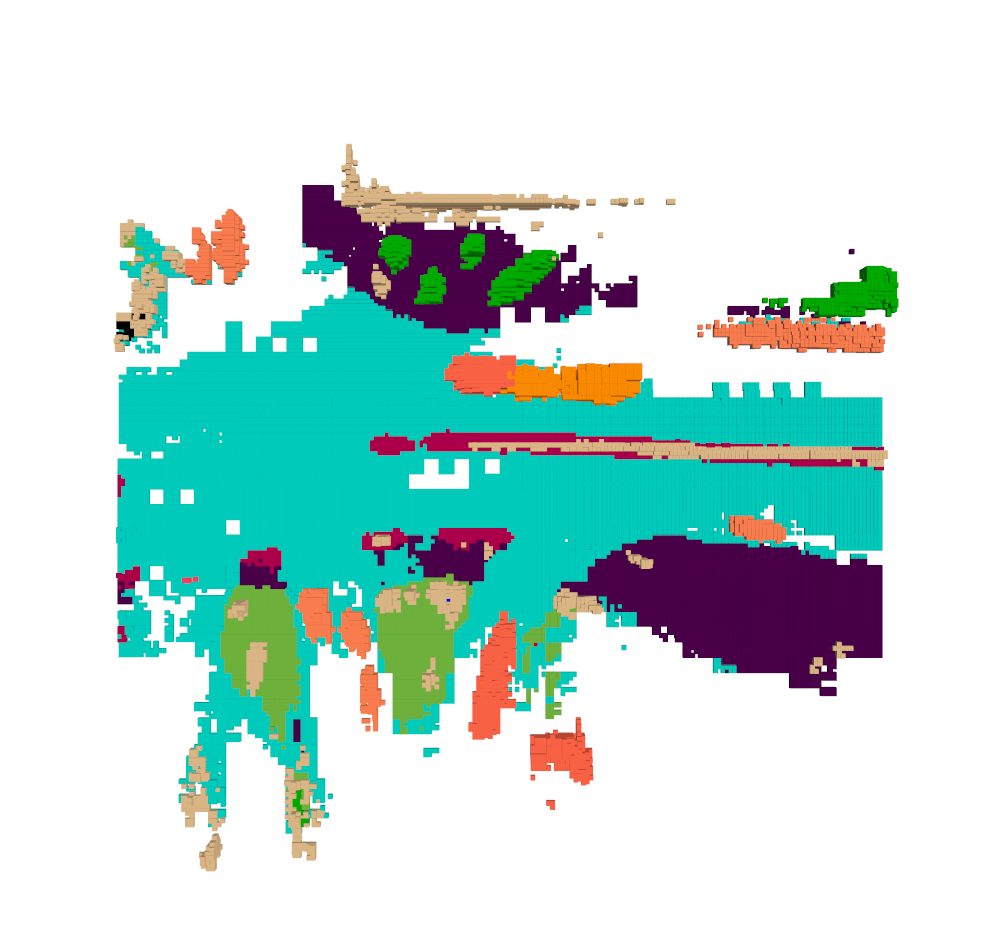}
    \includegraphics[trim={0pt 0pt 0pt 0pt},clip, width=0.24\textwidth]{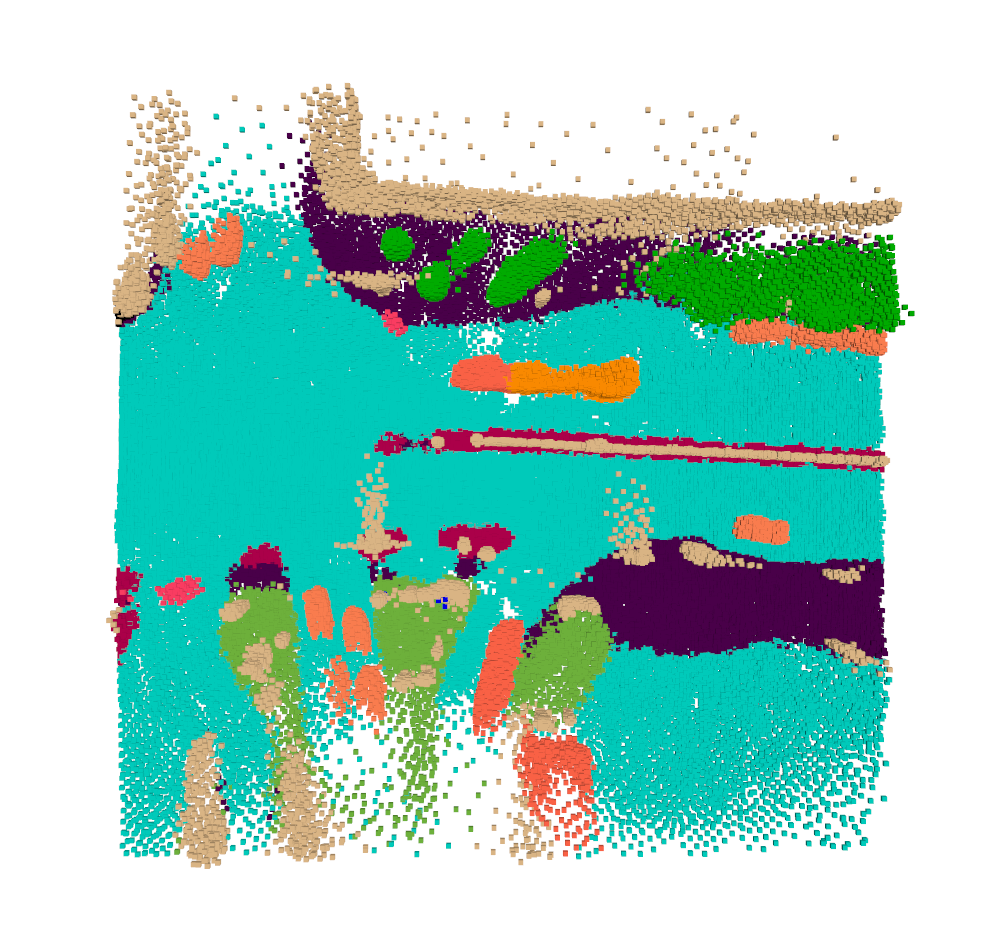}
    \includegraphics[trim={0pt 0pt 0pt 0pt},clip, width=0.24\textwidth]{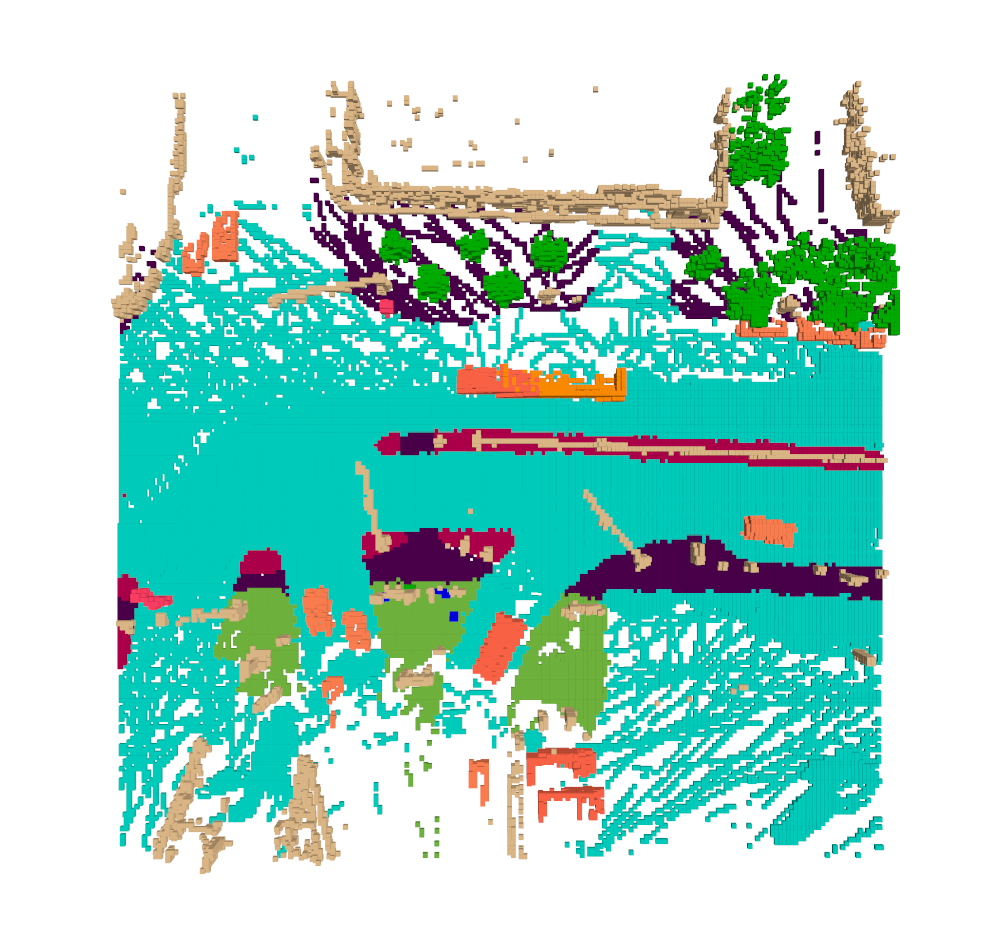}

    \includegraphics[trim={0pt 0pt 0pt 0pt},clip, width=0.24\textwidth]{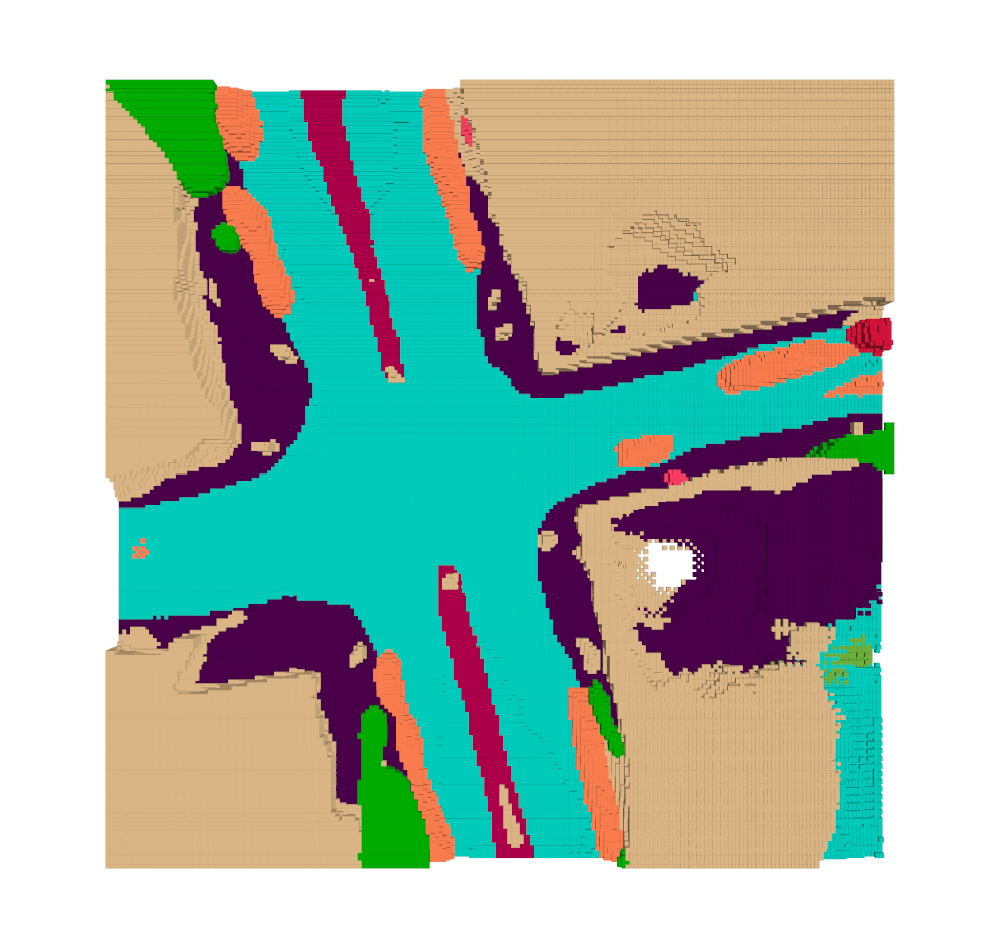}
    \includegraphics[trim={0pt 0pt 0pt 0pt},clip, width=0.24\textwidth]{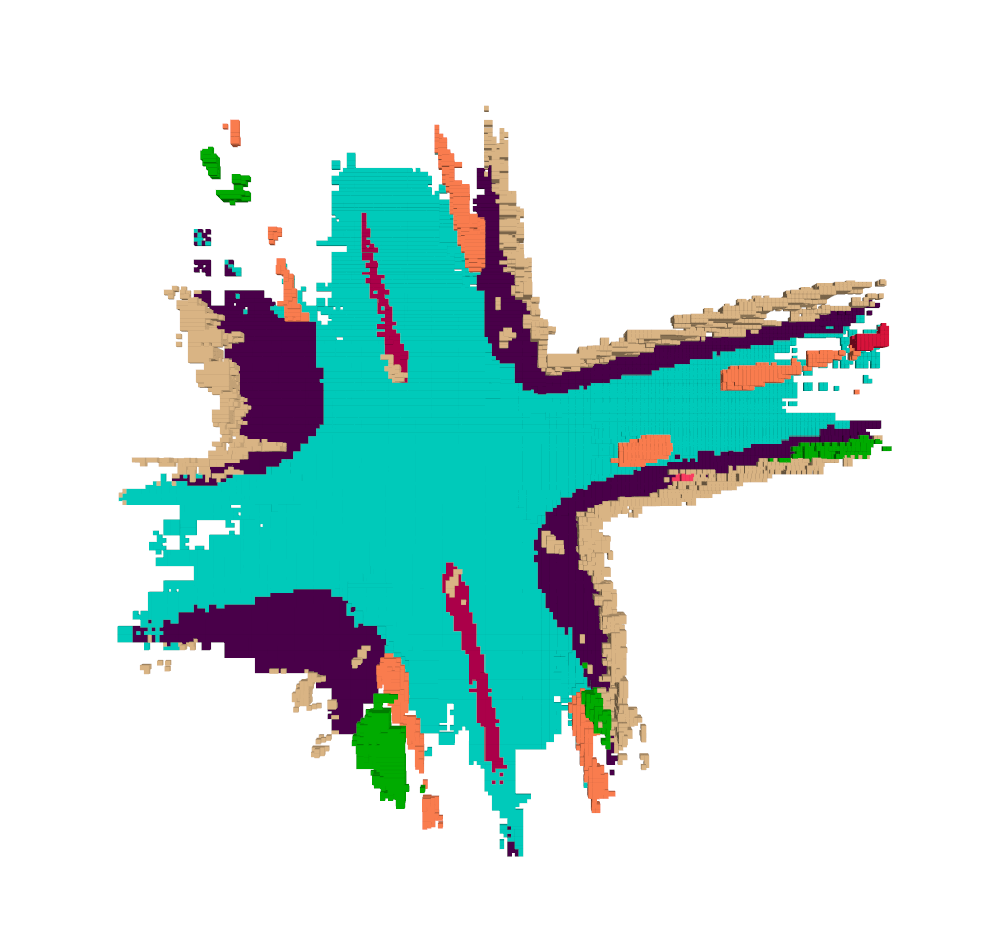}
    \includegraphics[trim={0pt 0pt 0pt 0pt},clip, width=0.24\textwidth]{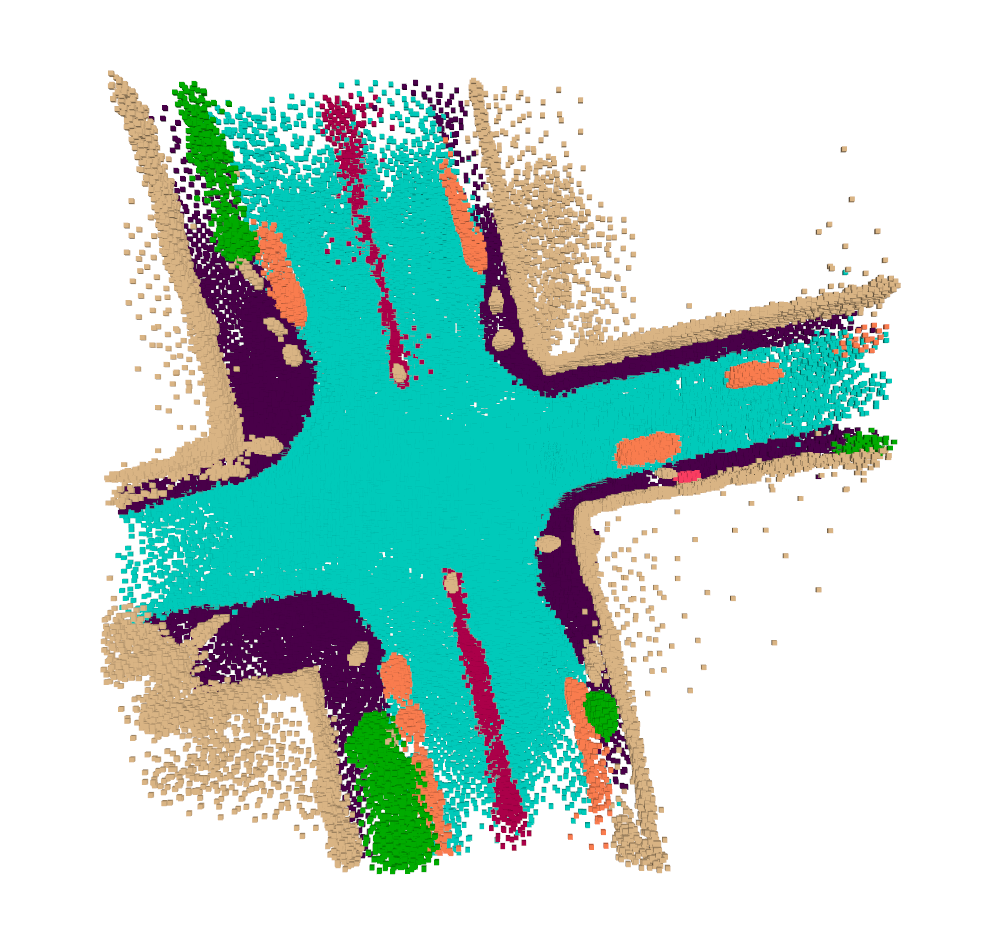}
    \includegraphics[trim={0pt 0pt 0pt 0pt},clip, width=0.24\textwidth]{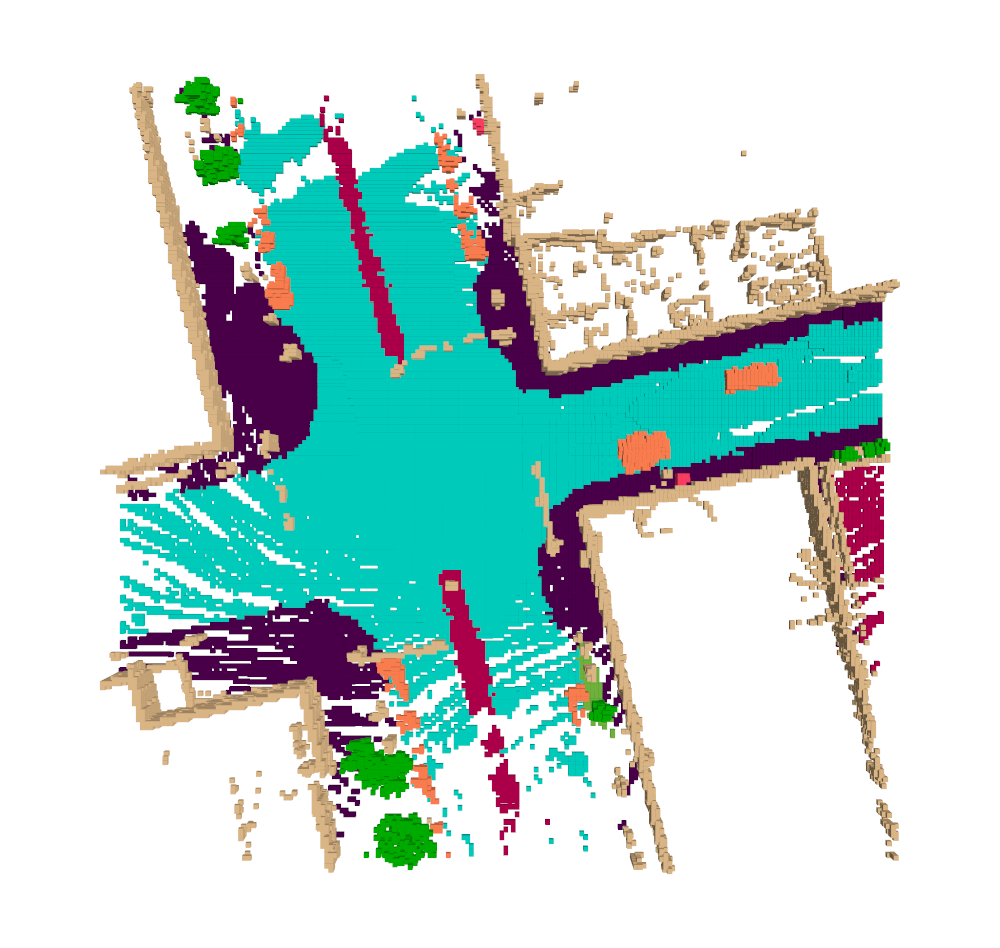}

    \includegraphics[trim={0pt 0pt 0pt 0pt},clip, width=0.24\textwidth]{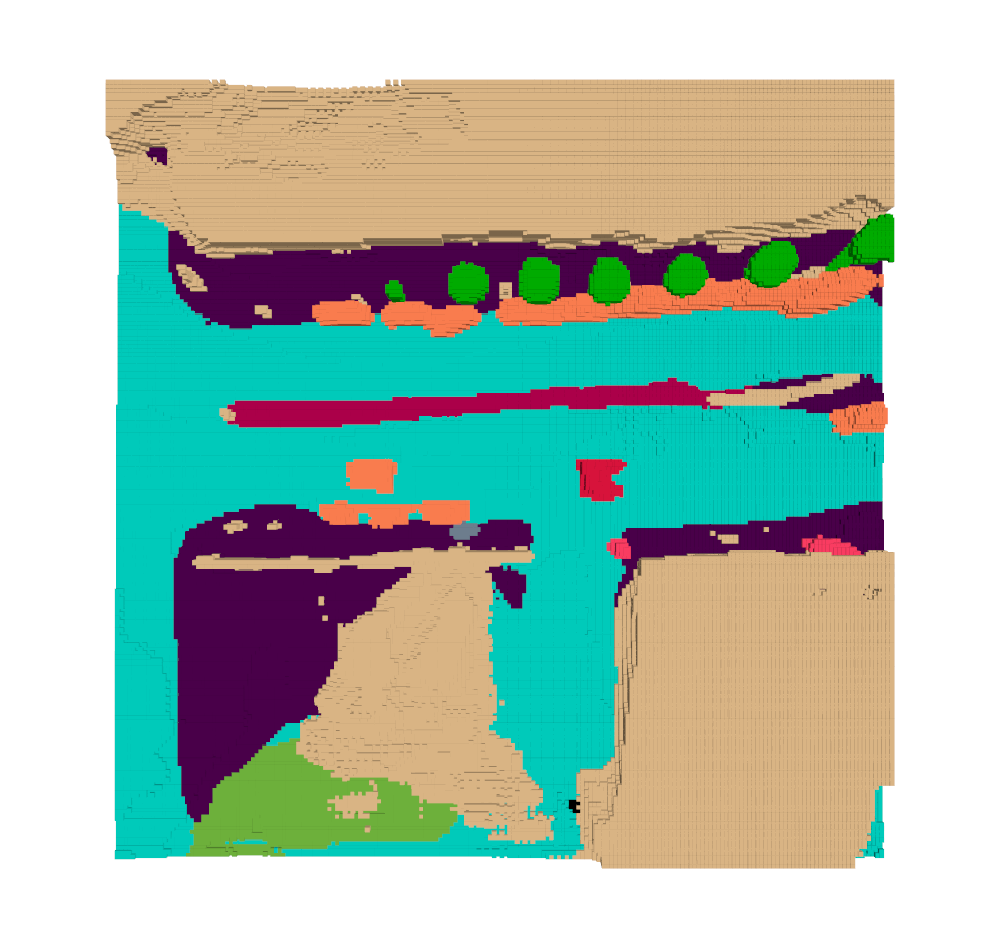}
    \includegraphics[trim={0pt 0pt 0pt 0pt},clip, width=0.24\textwidth]{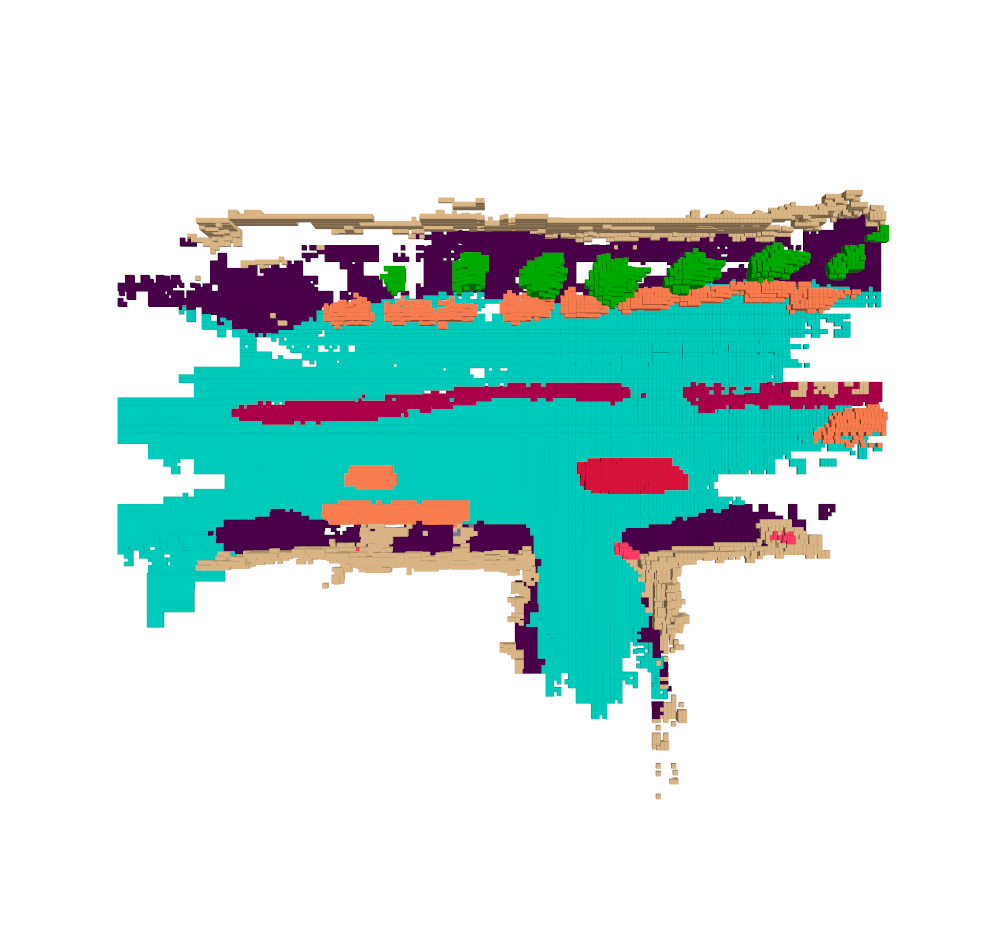}
    \includegraphics[trim={0pt 0pt 0pt 0pt},clip, width=0.24\textwidth]{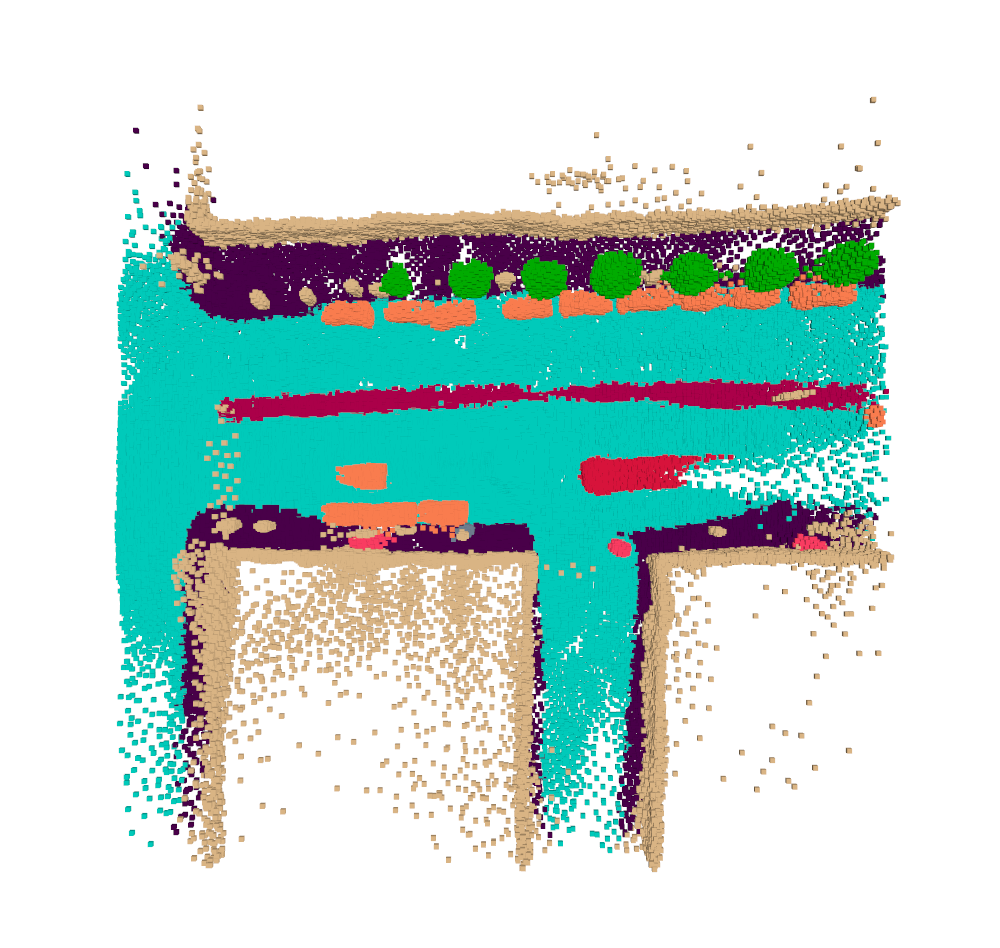}
    \includegraphics[trim={0pt 0pt 0pt 0pt},clip, width=0.24\textwidth]{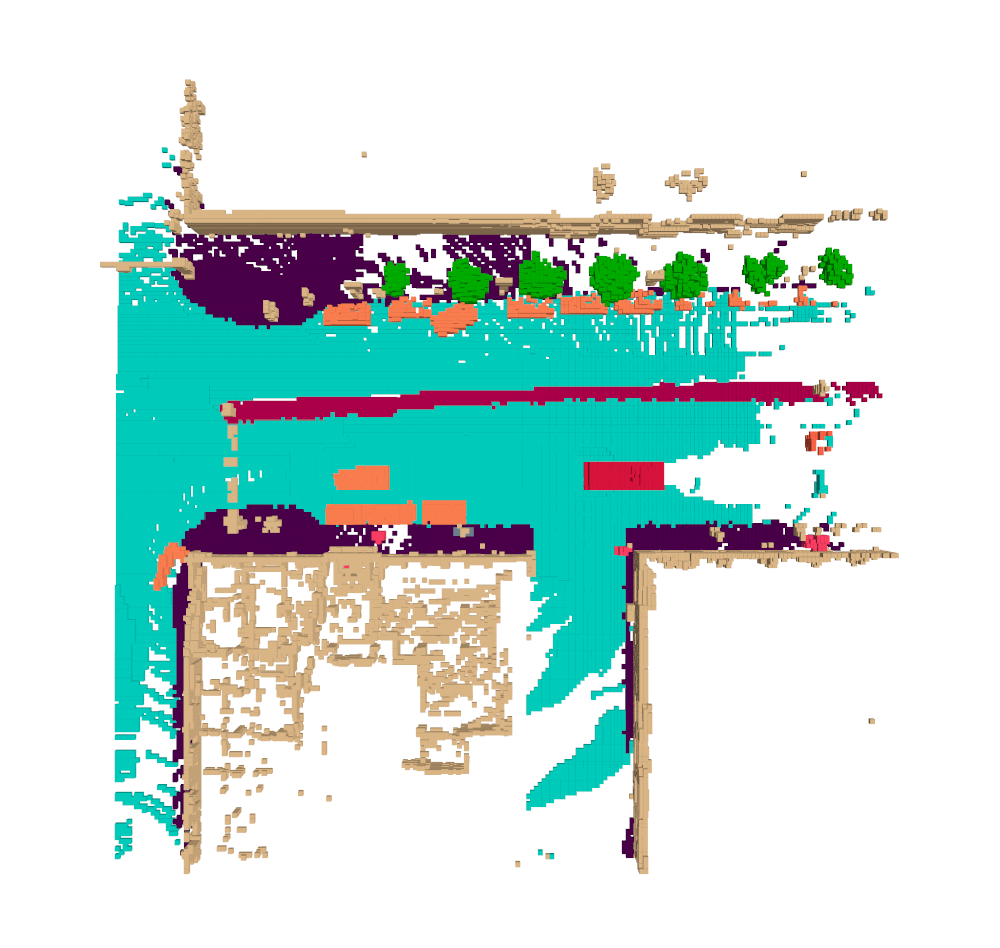}

    \includegraphics[trim={0pt 0pt 0pt 0pt},clip, width=0.24\textwidth]{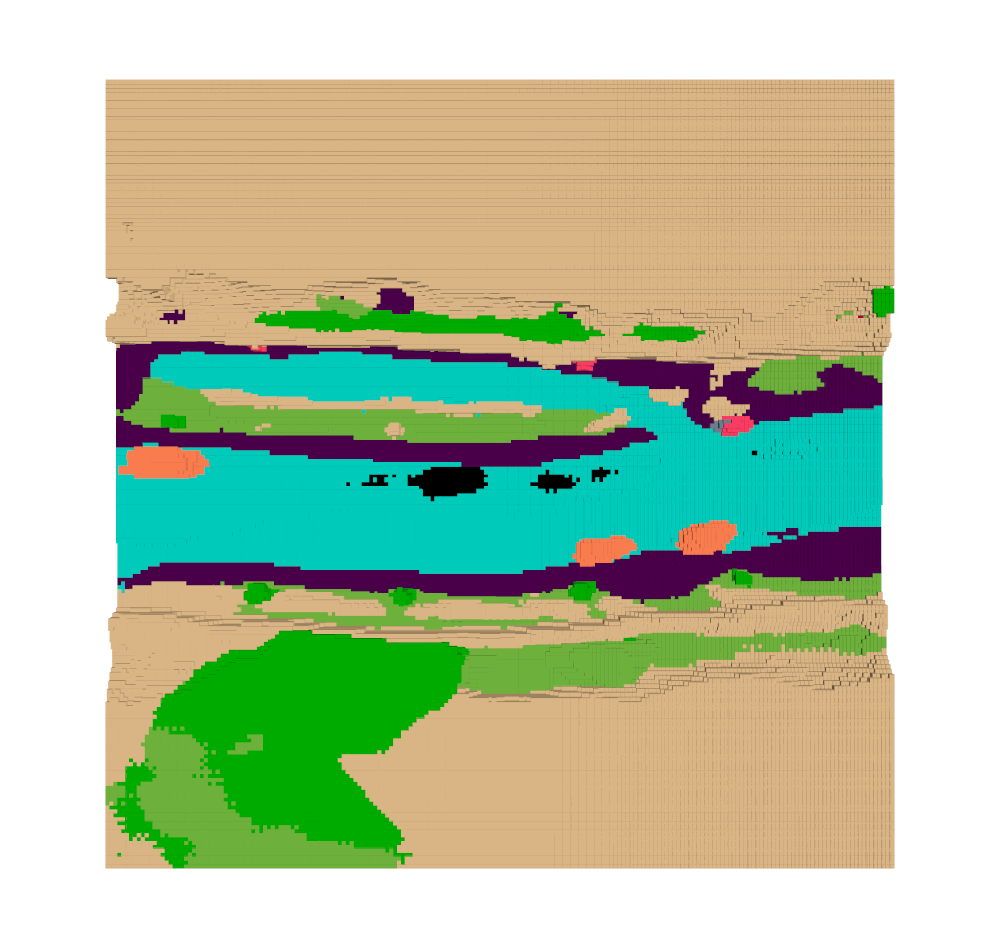}
    \includegraphics[trim={0pt 0pt 0pt 0pt},clip, width=0.24\textwidth]{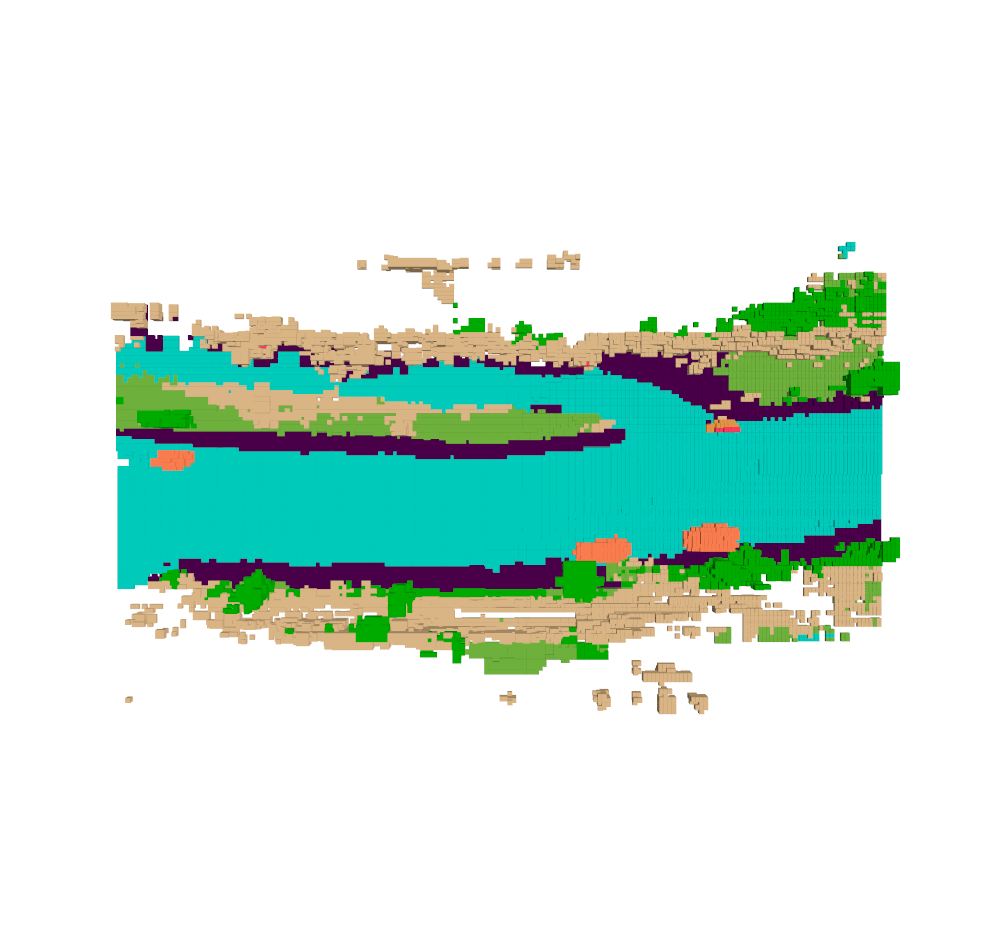}
    \includegraphics[trim={0pt 0pt 0pt 0pt},clip, width=0.24\textwidth]{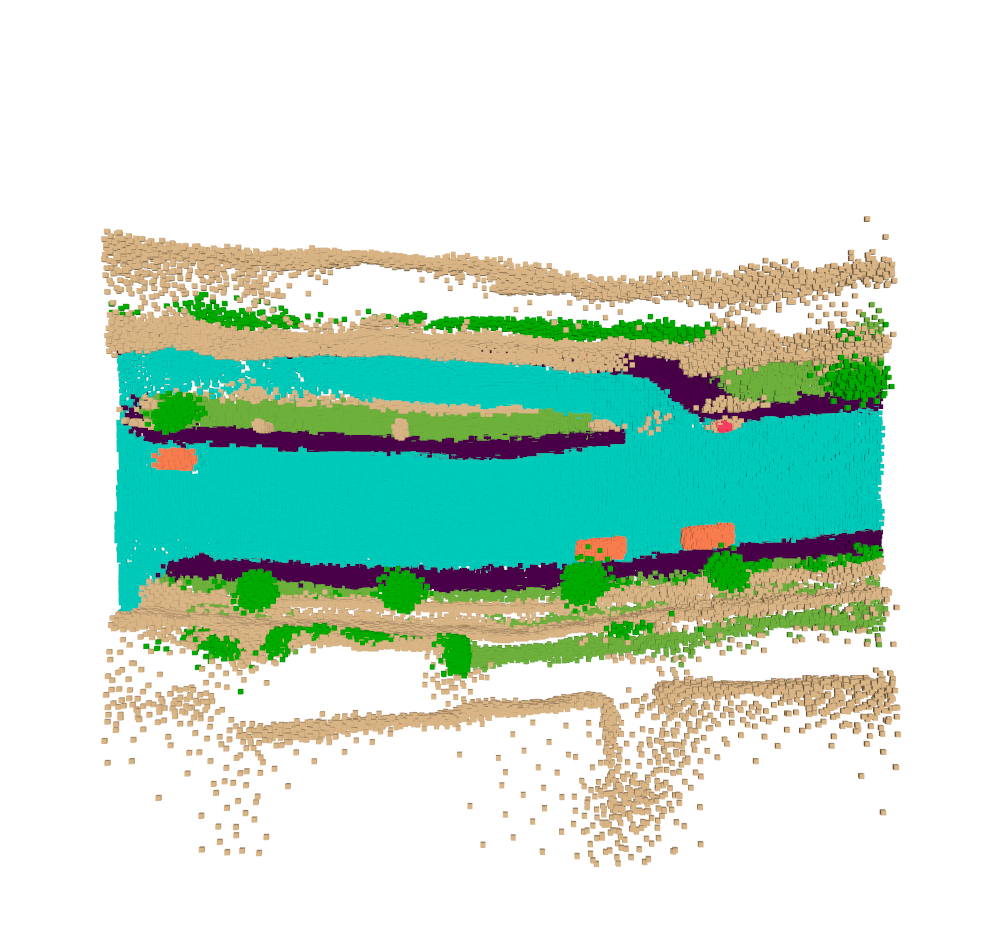}
    \includegraphics[trim={0pt 0pt 0pt 0pt},clip, width=0.24\textwidth]{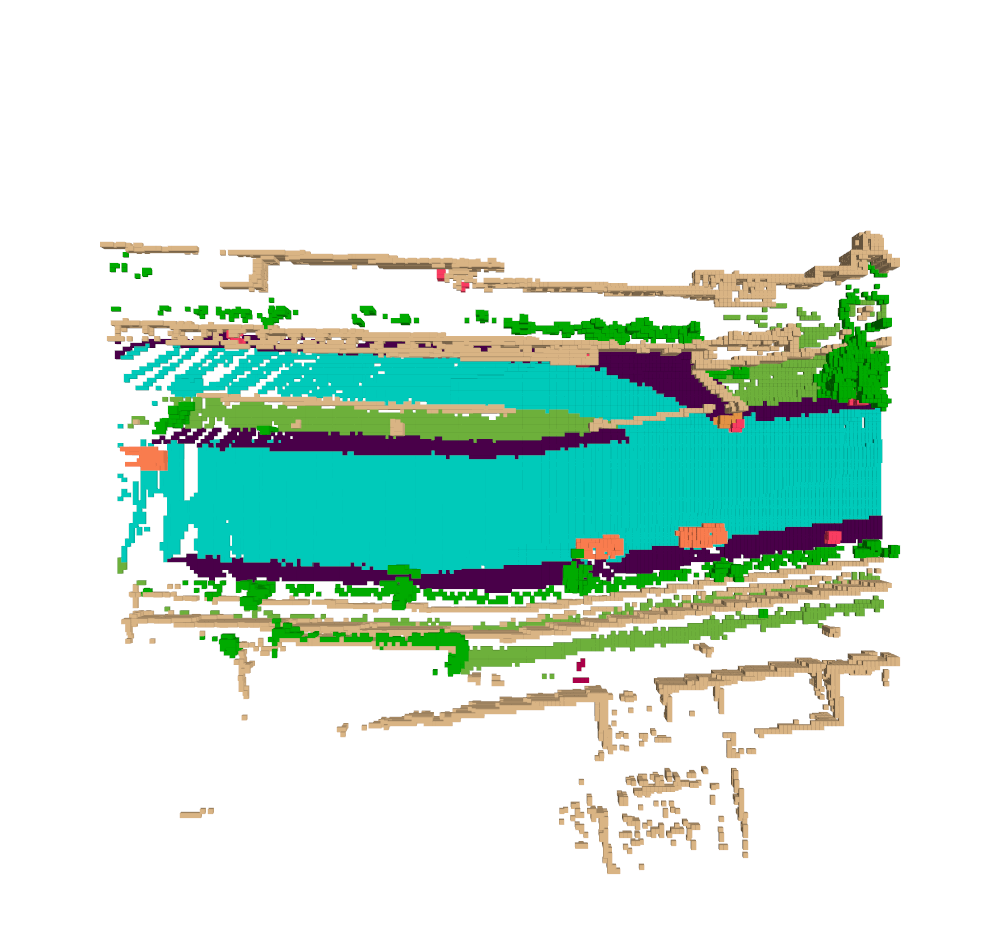}
    \vspace{-5pt}
    
    \caption{Visualizations of occupancy predicted by FB-Occ, SparseOcc and the proposed \Mymth{}.}
    \label{fig:vis_comp}
\end{figure*}

In \cref{fig:vis_comp}, We further provide more visualizations of occupancy predicted by FB-Occ, SparseOcc, and proposed \Mymth{}.
As shown in \cref{fig:main_compare} and \cref{fig:vis_comp}, a common \Mymth{} failure mode is the prediction of scattered and discontinuous surfaces at long distances.
Another is the presence of holes in predicted driving surface,  a phenomenon also observed in SparseOcc due to the sparsity properties.

\end{document}